\documentclass{article}

\usepackage[preprint,nonatbib]{neurips_2022}

\usepackage[square,numbers]{natbib}

\usepackage[utf8]{inputenc} %
\usepackage[T1]{fontenc}    %
\usepackage{hyperref}       %
\usepackage{url}            %
\usepackage{booktabs}       %
\usepackage{amsfonts}       %
\usepackage{nicefrac}       %
\usepackage{microtype}      %
\usepackage{xcolor}         %
\usepackage{floatrow}
\usepackage{graphicx}
\usepackage{subcaption}
\usepackage{wrapfig}

\usepackage{color}
\definecolor{darkgreen}{rgb}{0,0.5,0}
\definecolor{purple}{rgb}{1,0,1}
\newcommand{\kibitz}[2]{\ifnum\Comments=1\textcolor{#1}{#2}\fi}

\title{Fine-tuning language models to find agreement among humans with diverse preferences}

\author{Michiel A. Bakker\thanks{Authors contributed equally to this work}\\DeepMind\\\texttt{miba@deepmind.com} \And Martin J. Chadwick\footnotemark[1]\\DeepMind\\\texttt{martin@deepmind.com} \And Hannah R. Sheahan\footnotemark[1]\\DeepMind\\\texttt{hsheahan@deepmind.com} \And Michael Henry Tessler\\DeepMind\\\texttt{tesslerm@deepmind.com} \And Lucy Campbell-Gillingham\\DeepMind\\\texttt{lcgillingham@deepmind.com} \And Jan Balaguer\\DeepMind\\\texttt{jua@deepmind.com} \And Nat McAleese\\DeepMind\\\texttt{nmca@deepmind.com} \And Amelia Glaese\\DeepMind\\\texttt{glamia@deepmind.com} \And John Aslanides\\DeepMind\\\texttt{jaslanides@deepmind.com} \And Matthew M. Botvinick\\DeepMind\\ University College London\\\texttt{botvinick@deepmind.com} \And Christopher Summerfield\\DeepMind \\ University of Oxford\\\texttt{csummerfield@deepmind.com}}

\begin{document}

\maketitle

\begin{abstract}
Recent work in large language modeling (LLMs) has used fine-tuning to align outputs with the preferences of a prototypical user. This work assumes that human preferences are static and homogeneous across individuals, so that aligning to a a single ``generic'' user will confer more general alignment. Here, we embrace the heterogeneity of human preferences to consider a different challenge: how might a machine help people with diverse views find agreement? We fine-tune a 70 billion parameter LLM to generate statements that maximize the expected approval for a group of people with potentially diverse opinions. Human participants provide written opinions on thousands of questions touching on moral and political issues (e.g., ``should we raise taxes on the rich?''), and rate the LLM's generated candidate consensus statements for agreement and quality. A reward model is then trained to predict individual preferences, enabling it to quantify and rank consensus statements in terms of their appeal to the overall group, defined according to different aggregation (social welfare) functions. The model produces consensus statements that are preferred by human users over those from prompted LLMs ($>70\%$) and significantly outperforms a tight fine-tuned baseline that lacks the final ranking step. Further, our best model's consensus statements are preferred over the best human-generated opinions ($>65\%$). We find that when we silently constructed consensus statements from only a subset of group members, those who were excluded were more likely to dissent, revealing the sensitivity of the consensus to individual contributions. These results highlight the potential to use LLMs to help groups of humans align their values with one another.
\end{abstract}

\section{Introduction}

Modern large-scale transformer-based language models have revolutionized the capacity of AI systems to perform complex natural language processing tasks including reading comprehension, common sense reasoning, and fluent language generation \cite{brown2020language,rae2021scaling,chowdhery2022palm,hoffmann2022training}. 
A key challenge in language modelling is to ensure that the generated text is helpful, legitimate, and aligned with human values \cite{kenton2021alignment,weidinger2021ethical,askell2021general,gabriel2020artificial}. One popular approach is to recruit human participants to rate or compare candidate model outputs, providing feedback to the model about its performance on tasks like summarisation, instruction-following, and question answering \cite{bai2022training,ouyang2022training,stiennon2020learning,ziegler2019fine,menick2022teaching,Glaese2022ImprovingAO}. 
For large models, this approach improves performance on datasets specifically designed to test alignment (e.g., the HHH dataset), without decreasing their overall language competencies \cite{askell2021general}.  

Though powerful and general, extant methods for fine-tuning language models from human preferences treat these preferences as if they were homogeneous and static. 
This assumption is reasonable for a task such as article summarisation, where there is an objectively defined ground truth (i.e., facts in the article that the model must summarise).
However, for a wide variety of social problems that humans solve themselves using language (e.g., social coordination and group decision making), we cannot assume that people all share the same values.
A key case study in alignment of diverse preferences is consensus formation.
Consensus is commonly defined as the agreement of a large fraction of a social group about a particular topic or course of action. It is both a prerequisite for cooperation and a key pillar of the democratic process.
Finding consensus for humans is not easy, and technology often exacerbates political division rather than fostering reconciliation among those with divergent opinions \cite{aral2020hype}.
Large language models achieve surprising fluency and sensitivity to homogeneous preferences, but their ability to help people find agreement has not yet been tested.

Here, we investigate the use of large language models (LLMs) to aid humans in collectively producing written opinions that maximize approval rates among users.
Specifically, we create a corpus of thousands of questions concerning political issues relevant in the United Kingdom, about which reasonable, well-informed people might legitimately disagree (for example, ``Should we tax unhealthy foods and sugary drinks?'', or ``Should we re-nationalise the railways?''). 
We recruit groups of human participants to write out their opinions about these questions.
We then fine-tune a 70 billion parameter language model (\textit{Chinchilla} \cite{hoffmann2022training}) to produce candidate consensus statements that small groups of participants would be likely to endorse (see Table~\ref{table:example}), guided by an underlying family of social welfare functions.
Critically, in our work, the language model is not trained to adopt a particular opinion or persuade others of any one view. Rather, it is trained to produce consensus candidates based on the opinions contributed by the human group.
We find that our particular data collection and training pipeline results in a model that generates statements that are preferred more strongly than a number of high-performing baselines, including individual human opinions. The statements generated by the model reflect the underlying opinions that are contributed by the users.
This work opens up new possibilities for LLMs to help people find common ground in service of collective decision making.

\begin{table}[t]
    \caption{A candidate consensus statement generated by our SFT-Utilitarian model in response to a question and corresponding opinions from the test set. The example was chosen randomly from a set of consensus statements that included 3 opinions and that was in response to a divisive question (i.e., questions for which there was some disagreement). Other examples can be found in the Tables A6-7.}
    \label{table:example}
    \centering
    \footnotesize
    \begin{tabular}{@{}p{0.77in}p{\dimexpr \linewidth-2\tabcolsep-0.77in}@{}}
    \toprule
    \textbf{\normalsize Question} & Should we lower the speed limit on roads?\\
    \midrule \midrule
    \textbf{\normalsize Opinion 1} \newline \emph{participant} & The higher speed limits are ok and need no change. Some of the lower limits, like 30mph in built up areas, should be reduced and better enforced to reduce accidents. All in all the higher limits would not have much impact if reduced. \\
    \midrule
    \textbf{\normalsize Opinion 2} \newline \emph{participant} &Yes. Driving faster causes higher pollution and does not significantly shorten the average length of journeys. It would increase the fuel efficiency of those driving and prevent the distribution of microscopic particles of rubber from tyre wear into the environment. The number of road fatalities would reduce too. \\
    \midrule
    \textbf{\normalsize Opinion 3} \newline \emph{participant} & I think the speed limits are mainly fine the way they are.  In our area speed limits are reduced around busy areas such as schools and parks which is a good thing. \\
    \midrule \midrule
    \textbf{\normalsize Consensus} \newline \emph{SFT-Utilitarian} & We believe that speed limits are currently appropriate.  However, we feel that there is a need to enforce speed limits more strictly, particularly in areas with a high density of pedestrians such as near schools.  We also feel that there is a need to educate people more about the effects of driving too fast, such as the effects on fuel efficiency and pollution.\\
    \bottomrule
    \end{tabular}
\end{table}

\section{Related Work}

\textbf{Aligning LLMs with human preferences}. Several papers have reported the successful training of AI systems directly from human preferences (e.g., training deep RL agents to play Atari \cite{christiano2017deep,ibarz2018reward}). In the domain of language modelling, human preferences have been used to fine-tune models to perform specific tasks, such as summarisation or translation \cite{ziegler2019fine,stiennon2020learning,bai2022training}, to provide inline citations supporting their claims \cite{nakano2021webgpt,menick2022teaching,thoppilan2022lamda}, or for generic instruction following \cite{ouyang2022training}.  Our pipeline shares several design similarities with these works (e.g., the reward modelling approach). A key point of distinction, however, stems from the source of legitimacy in which statements produced by the language model are ostensibly grounded. For these related works, external sources that are generally agreed to be truthful and unbiased are the source of legitimacy. In our work, by contrast, we optimise, through a combination of opinion-conditional reward modelling and group-level welfare maximisation, to produce statements that will generate agreement for a specific group.

\textbf{Summarization}. Our work builds upon attempts to build natural language processing models to provide high-quality summaries of text (e.g., \cite{chopra2016abstractive, see2017get}).
These efforts have redoubled recently with the advent of LLMs \cite{ziegler2019fine,stiennon2020learning,rae2021scaling}. Like consensus opinions, good summaries should be concise, informative and balanced passages of text. However, summaries and consensuses differ in their grounding: good summaries are based on accurate, fact-based information about the world (for which a single viewpoint is typically accepted) whereas a consensus is grounded in the opinions of the specific individuals in a group seeking to achieve agreement.
Our work specifically builds on work in \emph{opinion summarisation}, in which subjective opinions (typically reviews about products, restaurants, or movies) are summarized into a kind of meta-review (e.g., \cite{amplayo2019informative,suhara2020opiniondigest}). While similar in spirit to our work on consensus generation, extant work in opinion summarisation falls short of using actual human feedback and does not verify generated summarisations with the same individuals who provided their opinions, instead focusing on \emph{third-party} evaluations (e.g., does this summary make sense in light of these opinions?). By not engaging with the people that provided the original opinions, these projects do not pertain to alignment problems but instead are closer to other summarisation work.

\textbf{Collective Reasoning}.
Our work is relevant to a fast-growing interest in the use of technology, including machine learning methods, to promote human collective reason, including democratic deliberation \cite{landemore2020open,fishkin2009people}. More traditional ML approaches (e.g., data/opinion mining) make use of richly structured (e.g., graph-based) models to make sense of public discussions \cite{zhang2018characterizing} and to facilitate interactions in online communities \cite{gu2021case}. We seek to leverage recent breakthroughs in large-scale language modelling to facilitate public deliberation via consensus generation.

\section{Methods}
\begin{figure}
    \centering
    \includegraphics[width= \textwidth,trim={0 3.2cm 0 0},clip]{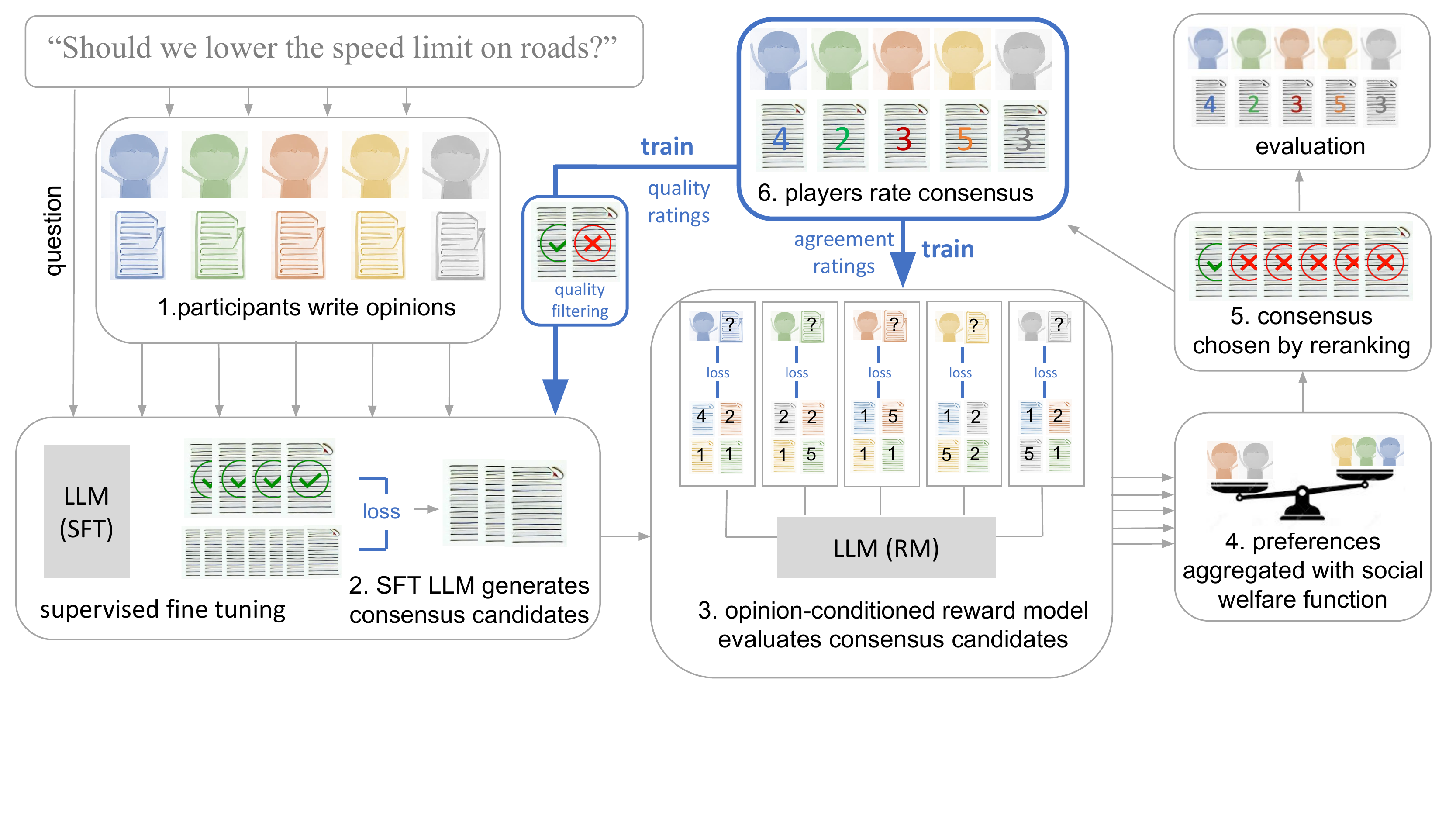}
    \caption{Overview of the data collection procedure. The evaluation pipeline proceeded in six steps. (1) Human participants, sorted into small groups ($n \in \{3,4,5\}$), each wrote a short paragraph stating their opinion about a political question (e.g., ``should we lower the speed limit on roads?''). (2) These opinions, together with the question, were passed to a prompted pre-trained LLM (or, a fine-tuned LLM on later rounds) via the prompt, which generated consensus candidates. (3) Pairs of participant opinions and candidate consensus statements were passed into a reward model, which estimated the degree to which each participant would agree with a candidate consensus. (4) For each consensus candidate, the set of predicted individual preferences were aggregated with a social welfare function. (5) From a batch of consensus candidates, the one that maximised welfare was selected for human evaluation. (6) Participants then rated this consensus candidate, together with candidates generated in other batches or conditions, on a 7-point agreement scale. Quality ratings were used to filter the data for later fine-tuning and agreement ratings for training the reward model and for evaluation.}
    \label{fig:pipeline}
\end{figure}

We created a large data set of debate questions and built a customized environment and pipeline that allowed us to collect human opinions and fine-tune our models in an iterative loop (Figure~\ref{fig:pipeline}).

\subsection{Generating debate questions}

We generate questions using a prompted, 70 billion parameter pre-trained LLM (\textit{Chinchilla} \cite{hoffmann2022training}). We seeded the process with 152 hand-written questions on issues of contemporary debate. Most were policy questions of the form "Should we...?" or "Should the government...?" that are relevant for the UK-based participants that took part in our human evaluation. We use these 152 seed questions to artificially generate a total of 3500 debate questions. For each debate question, we prompt \textit{Chinchilla} with a sample of 10 seed questions. The model generates a new question, which, if unique, we add to our total set. 
We manually check questions and filter out any that we consider likely to elicit extremist views or discriminatory language from the users.
 
This results in 2922 questions which we use to create a training set and two test question sets by clustering according to topic. We first embed each question using a Universal Sentence Encoder and cluster the questions into 110 topics using k-means clustering \cite{cer2018universal}. There is, for example, a cluster of 25 questions that corresponds to questions on food taxes such as ``Should there be a tax on junk food?'' and ``Should we remove all tax on food and groceries?''. We then split the clusters into two groups, and set some question-clusters aside for an out-of-distribution hold out set (n = 302). The remaining questions were sorted into a training (n = 2320 questions) and a within-distribution hold out set (n = 300). See Appendix A for more information on question generation, filtering and clustering.

\subsection{Data collection and environment design}\label{sec:data_collection}

On each round, groups of UK-based participants ($n=3211$ organized into $746$ groups, combined across training and eval) viewed a question, and wrote their opinion by typing freely into a text box in our custom online application. 
The opinions of a group were put into a prompt that was provided to one or more LLMs to generate candidate consensus statements. 
Data was collected using groups of four or five participants, though candidate consensus statements were sometimes generated based on a subset of participants' opinions (Section \ref{subsec:opinionexclusion}). Consensus candidates were then presented individually to participants in a random order, and each participant rated them along two dimensions, quality and agreement, using 7-point Likert scales. Participants rated all candidate consensus statements twice, to allow us to measure intra-rater reliability.

Additionally, before providing their own opinion or viewing candidate consensus statements, participants provided an agreement rating on a ''position statement'', which is a version of the question stated declaratively (e.g., ``We should...''). This allowed us to measure baseline disagreement among the group.
Each training or evaluation session took between 45 minutes and an hour. As our explicit goal is to train and evaluate on diverse opinions, we recruited a new set of participants for each data collection session, rather than use the same participants repeatedly. The full details of our study design, including compensation rates, were reviewed by our independent ethical review committee. All participants provided informed consent prior to completing tasks and were reimbursed for their time. It is our policy that researchers must pay workers/participants at least the living wage for their location. For this study participants were paid an average compensation rate of £15 per hour (the total cost of the study was approximately £46,000). No personally identifiable information was collected as part of this research, nor was any offensive content shown to participants. More detail on the environmental design and the recruiting and training processes can be found in Appendix B.

\subsection{Group alignment}\label{sec:group}

In line with previous work \cite{ziegler2019fine,stiennon2020learning,ouyang2022training}, we use reward models to predict whether generated statements will be preferred by participants. Given our focus on diverse preferences, we train a reward model that predicts agreement conditional on a person’s own opinion. Given a group of people with corresponding opinions, we can then use this model to generate, for each person, a score that predicts how likely they are to agree with a given statement.

Having estimated each individual’s expected agreement, we aggregate these scores to predict the extent to which a candidate consensus will be preferred by the group. For simplicity, we assume that preferences are cardinal and comparable between participants, and we aggregate them using a cardinal Social Welfare Function (SWF). SWFs are used in the field of welfare economics to map a set of numeric individual utitilities to collective welfare \cite{keeney1975group}. The goal of the model is then to generate a consensus that maximizes the group welfare given the set of opinions from the group. Any SWF that satisfies six desirable axioms belongs to a one-parameter family of isoelastic social welfare functions \cite{moulin2004fair}
\begin{equation}
    W_{\alpha}(u_1,\dots, u_n) =  
        \left\{
    \begin{array}{ll}
      \left[\frac{1}{n} \sum_{i=1}^n  u_i^{1-\alpha} \right]^{\frac{1}{1-\alpha}} & \mbox{if $\alpha\ge 0$, $\alpha \ne 1$} \\
      \\
      \sqrt[n]{\prod_{i=1}^n u_i} & \mbox{if $\alpha = 1$} 
    \end{array}
  \right.    
\end{equation}
where $u_i$ is the utility of person $i$ and $\alpha$ is the degree of inequality aversion. At one extreme, for $\alpha=0$, the SWF corresponds to max-mean or Utilitarian, which computes the mean expected agreement across the group. On the other extreme, for $\alpha=\infty$, the SWF corresponds to max-min or Rawlsian, which maximises the expected agreement for the most dissenting group member. Rawls argued in favour of the welfare function that yields the most desirable condition for the worst-off member of the group \cite{rawls_1971}. In the Results section we use a Utilitarian social welfare function for ease of interpretation but we present results on Rawlsian and Bernoulli-Nash (max-product, $\alpha=1$), in Appendix D.5. 

\subsection{Training}

Our training pipeline largely follows previous reports that use human feedback to fine-tune large language models \cite{ziegler2019fine}.\footnote{Supervised fine-tuning with demonstrations from expert human consensus-writers (with access to the opinions) could be a viable alternative. However, previous work on summarisation has shown that learning from demonstrations can result in the model prioritising the fine-tuning objective (maximising the likelihood of a demonstration, including low-quality examples) over the true objective (in this work maximising welfare) \cite{stiennon2020learning}.} Thus, we use a supervised fine-tuned LLM to generate $N$ consensus candidates ($N=16$), which are then reranked by a reward model according to their expected social welfare. We then select the statement that maximizes the welfare. We denote these models SFT-@SWF where @SWF corresponds to welfare function. 

The reranking approach gives us flexibility over the social welfare function during deployment.\footnote{Recent work has shown that reranking can perform on-par or better than directly optimizing a model to maximize human preferences using reinforcement learning \cite{menick2022teaching,thoppilan2022lamda}.} To ensure that the model generalizes to different social welfare functions on the Utilitarian-Rawlsian axis, we sample the inequality aversion parameter $\alpha$ during training time from a log-normal distribution. During training, we start with the 70 billion parameter pretrained Chinchilla model and iterate twice over the following training steps before evaluating our final models. 

\textbf{Step 1 - Generate consensus candidates and have them rated by humans} Participants provide written opinions in response to a question. The fine-tuned language model takes in the question and these opinions as part of its prompt and generates a set of consensus candidates (on the first iteration, we bootstrap first with zero-shot prompting and then with few-shot prompting of the base \textit{Chinchilla} model). To ensure that our dataset contains data on a variable number of opinions (between 3 and 5), each time a statement was generated we silently omitted 0, 1, or 2 of the participants’ opinions. For each unique number of opinions, we generate 16 candidates and select 2, each of which is ranked top-1 under a different $\alpha$. These 6 (3x2) highest ranked consensuses are then presented to participants who rate them for quality and agreement using two 7-point Likert scales. Note that, on the first iteration, we have no reward model for the reranking and selection scheme and thus simply generate $2$ candidates for each unique set of opinions. Across all training data collection runs, 1524 participants contributed to our training data. See Appendix B.1 for more details.

\textbf{Step 2 - Supervised fine-tuning (SFT) to improve quality} We fine-tune a pretrained \textit{Chinchilla} model on the consensus candidates that were rated as high quality (mean quality of 6 or higher on a 7-point Likert scale). The purpose of SFT is to familiarize the model with the prompt template and increase the candidate quality. We do not use agreement ratings to filter candidates as we aim to retain diversity in the kinds of stances expressed in the candidates so we can then use reranking to find the best candidate in terms of welfare\footnote{Our SFT approach is similar to RL via Expert Iteration \cite{anthony2017thinking}, alternating between (1) policy improvement, sampling from the current model and filtering these samples for quality, and (2) distillation, training a new model using these filtered samples.}. SFT training details including the prompt template and hyperparameters can be found in Appendix C.1.2. 

\textbf{Step 3 - Train a reward model (RM) to predict preferences} We train an RM to take in a question, opinion and a statement, and output a scalar "agreement" score. The score is a proxy for how likely an individual is to agree with a statement given their own opinion. To provide data for model training, each participant rates the six consensus candidates along a 7-point Likert score. Note that, as we generate candidates based on 3, 4, and 5 opinions, participants also rate candidates that were generated without taking their opinion into account. We map these six ratings to $6 \choose 2$ pairwise comparisons, remove the rating ties, and remove ratings from participants with low intra-rater reliability (see Appendix B.2.1). The RM is then trained to predict which statement out of a pair each user will prefer, conditional on the question and opinion, using standard cross-entropy loss. Note that the RM only conditions on one opinion, while the SFT conditions on all opinions. We warm-start the RM using a pretrained \textit{Chinchilla} model and add an extra final linear layer to predict the reward. The prompt template and further training details can be found in Appendix C.2. After training the reward model, we go back to step 1.

\section{Results}

We ran two human evaluations, comparing our model against both high-quality baselines and human-generated opinions. These evaluations used the same data collection platform and basic design (see  Section \ref{sec:data_collection}) and were run in separate sessions using the within-distribution questions and out-of-distribution questions. We ran an additional evaluation experiment to assess the model's sensitivity to different social welfare functions, but this did not reveal any reliable  differences in the average or minimum agreement rating under any of the welfare functions of theoretical interest (we discuss this null result further in Appendix D.5).

\begin{figure}
    \includegraphics[width=1\textwidth]{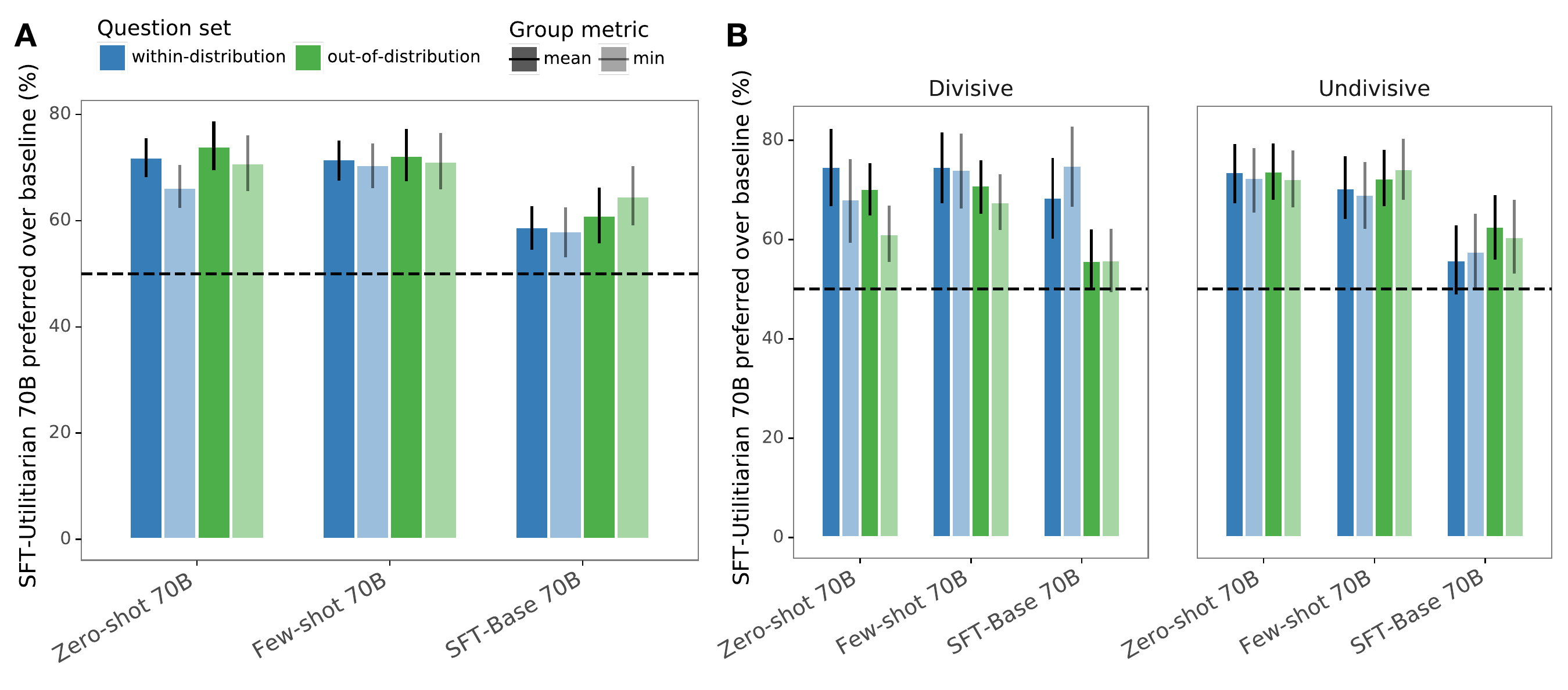}
    \caption{Win rates for comparing models constructed by pairwise comparison of Likert agreement ratings for candidate consensus statements (excluding ties) for within-distribution (blue) and out-of-distribution (green) question sets. Likert agreement ratings are aggregated within groups by either the mean (dark bars) or the minimum (light bars) agreement score. A: Win-rates for the SFT-Utilitarian model in comparison to baselines. B: Win-rates for the SFT-Utilitarian model broken down by whether or not the question was divisive in the group (see main text for details). Error-bars represent 95\% bootstrapped confidence intervals. }
    \label{fig:abblation_win_rates}
\end{figure}

\subsection{Preferences for consensus over baselines} 

We first test our main model against a set of high-quality baseline models. Our main model (\textbf{SFT-Utilitarian}) generates statements using a 70 billion parameter language model fine-tuned on high quality consensus statements. At inference time, we sample 16 statements based on the question and  opinions provided, and select the statement that maximizes predicted welfare under a Utilitarian (max-mean) aggregation function. The baseline models are:
\begin{itemize}
    \item \textbf{SFT-Base}. Our fine-tuned model but without the aggregation function selection process, sampling only one statement at inference time.
    \item \textbf{Few-shot}. A few-shot prompted \textit{Chinchilla} model. Each prompt contains three real examples composed of consensus statements with three, four and five opinions. Each set of examples is sampled from data collected using a zero-shot model, using a combination of quality and agreement criterion (see Appendix C.1.1 for more details).
    \item \textbf{Zero-shot} A prompted \textit{Chinchilla} model without examples.
\end{itemize}

Under each of the four model types, we generated two different candidates, all of which were presented to the participants for rating. In these evaluations against baseline models we examine and compare performance of the model under within-distribution questions (collected using groups of five participants; total $n=530$), and out-of-distribution questions (collected using groups of four participants; total $n=267$).

We found that the position statements on approximately 50\% of rounds (questions addressed by particular groups) were \emph{undivisive} (receiving either all agreement or all disagreement from the group, when examined in a binary fashion; see Appendix D.6); we perform a split of the dataset into \emph{divisive} and \emph{undivisive} questions for further analysis. Notably, the fact that 50\% of questions contained at least one dissenting participant (out of a group of up to five people) indicates that our population of participants do indeed have divergent opinions on many topics.
We first compare how likely a group is to prefer one policy over another. 
We do so using two metrics: the mean agreement across the group (consistent with the Utilitarian training objective of the model) and the minimum agreement in the group (consistent with a Rawlsian objective, not explicitly used in the model).\footnote{We additionally analyze the group agreement scores in terms of the median, which is more appropriate for an ordinal (Likert) scale, and is more robust to outliers in small samples, as we have here in our groups. Usage of the median results in the same substantive conclusions for the primary analyses and for clarity we report these results in Appendix D.3}
We compute these metrics for each statement and take the mean metric value across the statements generated under the same policy. 
To compute the win-rate, we compare these mean scores between two policies where the "winning" policy is the one with higher score. 
Under the group-mean agreement score, participants significantly prefer our main \emph{SFT-Utilitarian} model over all of the baselines (Figure~\ref{fig:abblation_win_rates}A), and strongly so in comparison to the two prompted LLMs.
Even more impressive, participants prefer the \emph{SFT-Utilitarian} model over all baselines for the group-minimum agreement score (a Rawlsian objective), indicating that our model is also more adept at increasing agreement among the strongest dissenters.
Furthermore, we see the human preference for our model's consensus statements is present for both more and less divisive questions (Figure~\ref{fig:abblation_win_rates}B), highlighting the model's utility in helping people find agreement even where opinions are divided.

In addition to the main baseline model comparisons detailed here, we also compared our main model against a smaller 1.4B parameter model ($n=224$). We found that training a smaller model with data generated from the larger model can be effective, 
as the \emph{SFT-Utilitarian-1.4B} outperforms the larger prompted models. However, the larger SFT-Utilitarian-70B is still superior, indicating that fine-tuning and size are both additive in this task. See Appendix D.1 for full details.

\begin{figure}
    \centering
    \includegraphics[width=\textwidth]{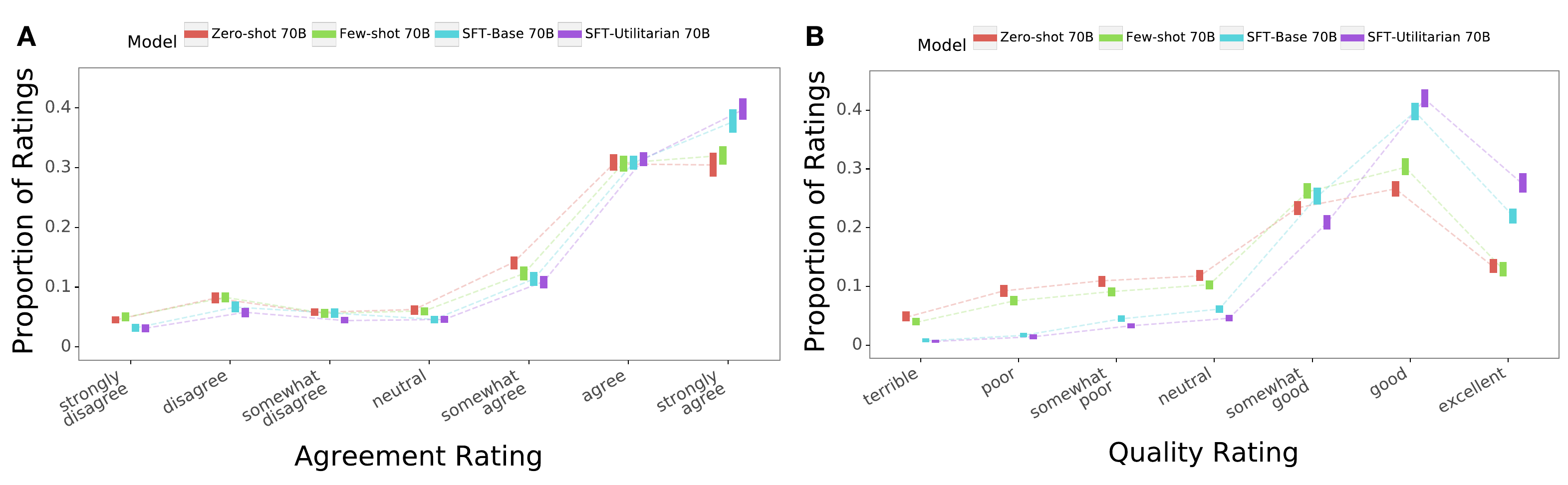}
    \caption{Distributions over Likert ratings for candidate consensus statements generated by the SFT-Utilitarian model and baseline models. A: Agreement ratings. B: Quality ratings. Error-bars represent 95\% bootstrapped confidence intervals. See Figure A11 for agreement scores broken down by question divisiveness.}
    \label{fig:Likerts}
\end{figure}

To further assess the reliability of the human preference for our \textit{SFT-Utilitarian} model's consensus generations over those of the ablated models, we constructed a maximal mixed-effects logistic generalized linear regression model \cite{liddell2018analyzing}, the standard in confirmatory hypothesis testing \cite{barr2013random}.  Consistent with our win-rate analysis, we found that participants agreed more strongly with the consensus candidates generated by the \emph{SFT-Utilitarian model} than those generated by the \emph{SFT-Base} ($\beta = 0.12$; SE = $0.047$; z = $2.53$; $p = 0.011$; Figure~\ref{fig:Likerts}A). Under this analysis, participants also preferred the \emph{SFT-Base}'s generations more so than those of the \emph{few-shot} model ($\beta = 0.40$; SE = $0.055$; z = $7.35$; $p < 0.001$), but exhibited no preference for the 
\textit{few-shot} model over the \textit{zero-shot} model ($\beta = 0.05$; SE = $0.06$; z = $0.88$; $p = 0.34$).
We observe that the quality of the model-generated statements also increases as a function of our training pipeline, with both supervised fine-tuning and reward modelling impacting participants' quality ratings (Figure~\ref{fig:Likerts}B).

To assess the performance of the SFT-Utilitarian model in more absolute terms, we examined the data for the approximately 50\% of rounds in which participants were on different sides of a position statement (i.e., the \emph{divisive} rounds). We found that, on these divisive rounds, 65.6\% [61.9, 69.3] of candidates generated by the SFT-Utilitarian model were less divisive than the initial position statements. Furthermore, on 40.8\% [35.4, 46.2] of rounds with a divisive position statement, a candidate consensus statement from the SFT-Utilitarian model achieved unanimous support (i.e., all participants somewhat agreed with the statement; see Appendix D.7 for full results and details). These results demonstrate not only that human participants preferred the consensus candidates generated by the SFT-Utilitarian model over those of the ablated models, but that the SFT-Utilitarian is demonstrably effective at helping a group of people with diverse preferences find points of agreement.

\subsection{Preferences for model candidates over human opinions}
\label{subsec:humanopinion}

Here we compare the performance of our \textit{SFT-Utilitarian} model against human-generated opinions. While the human opinions are not explicitly written with consensus in mind, the opinions are very high-quality, frequently containing well-reasoned justifications for their positions (see Tables A6-7 for examples).
Data was collected using groups of four participants ($n=189$ for within-distribution questions, $n=186$ for out-of-distribution questions). 
As with our previous evaluations, each participant wrote an opinion in response to the question, and these opinions were used to generate three candidates with our \textit{SFT-Utilitarian} model. The same participants were then shown these model-generated consensus candidates alongside the (anonymous) opinions of the other participants in their group. All participants rated both the consensus candidates and the opinions.

We perform two analyses: for each participant, we compare their mean agreement over the set of other participants' opinions to their mean agreement over the set of model candidates; we also compare their most preferred other opinion to their most preferred candidate.
We then compute the win rates over all participants and questions for each of these two metrics separately. The mean candidate score is preferred over the mean opinion score $78\%$ (95\%-bootstrapped CI: $[75\%, 80\%]$) of the time. Perhaps more impressively given the potential variance across the different opinions, even when we select the best-rated opinion, we find that our best-rated model candidate is still preferred $65\% [61\%, 69\%]$ of the time. Finally, we find that this difference in preference is larger for more divisive questions (win-rate of $66\% [61\%, 71\%]$) than for less divisive questions (win-rate of $63\% [57\%, 69\%]$).

\subsection{Opinion exclusion analysis}\label{subsec:opinionexclusion}
Our model's ability to generate statements that provoke higher agreement could be achieved by generating statements that are more likely to be preferred \emph{in general} but not tailored to the specific set of opinions of the small group. In order to assess this potential failure mode, we make use of the fact that our main evaluation dataset contains candidates that are based on subsets of opinions (3 or 4) from the full set of five provided by the participant group. If the model is making use of the specific opinions passed into it, then the agreement rates from participants whose opinions are excluded from the candidate generation process should be lower than those whose opinions are included. For each candidate in our evaluation set where fewer than 5 opinions were included, we compute the difference in median agreement between the inclusion group and the exclusion group.
This analysis reveals an average inclusion/exclusion Likert agreement difference of $0.47$ (CI = $[0.21, 0.73]$) for our \textit{SFT-Utilitarian} model. Quality ratings, by contrast, are not impacted by opinion exclusion (Mean$=0.10$, CI$=[-0.05, 0.25]$), and the difference between quality and agreement scores under this exclusion analysis is statistically significant ($t=2.94, p=0.0036$). This dissociation supports the conclusion that our model is successfully producing consensus candidates based on the specific set of opinions provided by the users, rather than producing generically preferred statements.

\subsection{Out-of-distribution generalisation}

So far, our evaluations are based on a question set that was unseen during training but that came from the same topic clusters of questions that were used during training. Next, we run separate human baseline experiments using the out-of-distribution question dataset, which has been specifically created so that the questions came from topic-space clusters that were never seen by our models during training. Despite the topic novelty, we find that that the \emph{SFT-Utilitarian} model's statements are preferred over all baselines to a degree that is numerically similar to that for the within-distribution questions (Figure~\ref{fig:abblation_win_rates}).
\emph{SFT-Utilitarian} model also continues to outperform the human opinions in the out-of-distribution question set at similar rates to the within-distribution question (mean candidate~vs.~mean human opinion: 76\% [74\%, 79\%]; max candidate~vs.~max human opinion: 59.5\% [56\%, 63\%]).
Overall, this additional set of evaluations demonstrates the capability of our model to generalize beyond its training distribution without apparent loss of performance.

\section{Discussion}
We fine-tune an LLM to take in a question and the written opinions of a human group, and generate a statement that maximises the agreement of that group. This work opens up new avenues for language modelling in which the goal is to accommodate a set of diverse preferences.

\subsection{Limitations}\label{sec:limitation}

\textbf{What makes a good consensus?} Consensus statements generated by our model are rated more highly than those produced under rival methods, including one baseline (SFT-Base) that mimics our approach in every respect, omitting only the reranking step. Moreover, the model is sensitive to the specific opinions provided in the prompt, because auxiliary analyses show that excluded participants offer lower agreement but similar quality ratings. However, we do not know exactly \emph{why} the statements are preferred. In particular, SFT-Utilitarian also receives a higher number of ``excellent quality'' ratings (see Figure~\ref{fig:Likerts}B) raising the possibility that its statements (whist tailored to the group) are also more generally sensible or well-written. Alternatively, people may use "quality" as a proxy for agreement, raising the additional risk that more confident or authoritative users will wield more influence over the consensus, potentially exacerbating power imbalances between different social groups.

\textbf{Social Welfare Functions.} We train the model under a distribution of SWFs. In the Results section we show that reranking statements using a Utilitarian aggregation function generates consensus statements that already take into account both minority and majority views (improving both the min and mean rating over baselines). Hence, when we compare the Utilitarian SWF to other more equitable welfare functions like Rawlsian and Bernoulli-Nash, we do not observe meaningful differences in the mean ratings across participants or that of the most dissenting (min) participant. We thus cannot conclude based on these results alone that aggregating using different SWFs actually results in different behavior. We report these results in Appendix D.5.

\textbf{Data collection} We collect human data from a crowd-sourcing platform. Because the debate questions are relevant to UK current affairs, we limit inclusion to UK participants. However, this curtails the diversity in our participant cohort, and limits the generalisability of our findings. It also raises the risk that consensus statements may unduly reflect the views or biases of the participant demographic that we have sampled, in addition to any biases that may arise during pretraining \cite{weidinger2021ethical, bender2021dangers}. We note that the benefits of our model are most pronounced at the upper end of the Likert scale, perhaps because homogeneity in our sample inflates baseline levels of agreement. We are developing additional recruitment methods to allow us to sample a more diverse group of people for future experiments.

\textbf{Scale} We limit data collection to small groups of four or five people. This choice is in part because of the technical limitation imposed by the prompt length of the model, which can handle only a handful of written opinions. There may, however, be important future use cases where aggregating over many thousands of opinions is necessary, requiring an architecture that is scale invariant. One approach that could be fruitful is to map each opinion to an embedding and then aggregate those embeddings directly. Related examples include the attention-based architecture in \cite{vinyals2015order} or the recursive approach developed for summarizing books in \cite{wu2021recursively}.

\subsection{Broader Impacts}\label{sec:broader_impacts}

\textbf{Misuse for persuasion} We did not train the language model to adopt a particular position or persuade others of a specific political view. Nevertheless, there is a risk that LLMs can be used for human persuasion, posing a risk in political discourse, the media, or in advertising. Political debate is already increasingly polarised, especially in Western societies. A system that is capable of persuading others to adopt a particular viewpoint could learn to present arguments in a manipulative or coercive manner, and mitigations for these potential harms play an important part in any research addressing this topic.

\textbf{Factuality} The language model we describe is not specifically fine-tuned to produce consensus opinions that are factually accurate. Thus, whilst manual review of consensus statements suggested that they were broadly accurate, there exists a risk that the consensus opinions that it produces could be
misleading or contain false information.

\textbf{Misrepresentation of consensus} 
Our work describes an AI that helps people find agreement in natural language. However, the consensus statement might not reflect the views of all the users. This raises important questions about how such a consensus is used. For example, a consensus might be presented as reflecting a unanimous view, misrepresenting the minority opinion, or use the consensus statement to justify otherwise unwarranted courses of action. It is important that users understand these caveats when interpreting the consensus.

\textbf{Opportunities} Nevertheless, despite these acknowledged risks, our research was conducted with societal benefit firmly in mind. The ultimate goal of our work is to provide a tool that can be used safely to help people find agreement. We focus on opinions about debate questions, but we can envisage a wider set of use cases, such as aggregation of online reviews into more helpful meta-reviews, systems for collective writing that automatically takes the preferences of different authors into account, and systems for collective decision making for organised groups. However, we note that considerable work is needed to understand the potential risks associated with AI consensus generation, and to find ways to ensure that model outputs are generated in a transparent and explainable way, before any such system can be deployed.

\section*{Acknowledgements}
We would first like to thank the human participants, whose high-quality opinions and ratings allowed us to train and evaluate our models. We would also like to thank Jonathan Uesato and Fan Yang for help with the LM infrastructure, Sarah Henderson, Richard Ives, Antonia Paterson and Jacklynn Stott for operational support, and Iason Gabriel, Will Hawkins, Geoffrey Irving, Raphael Koster, Angeliki Lazaridou, and Boxi Wu for helpful comments and suggestions.

\section*{Author Contributions}
\textbf{Experiments} were designed by: Michiel Bakker$^*$, Martin Chadwick$^*$, Hannah Sheahan$^*$, Christopher Summerfield, MH Tessler, Lucy Campbell-Gillingham

\textbf{The SFT and RM training schemes} were designed, built and executed by: Michiel Bakker, Nat McAleese, Hannah Sheahan

\textbf{The welfare-based reranking scheme} was designed and built by: Michiel Bakker$^*$, Martin Chadwick$^*$, Hannah Sheahan$^*$, Christopher Summerfield.

\textbf{The human data app and infrastructure} were designed and built by: Martin Chadwick$^*$, Hannah Sheahan$^*$, Jan Balaguer, Michiel Bakker

\textbf{The human data} was collected by:  Martin Chadwick$^*$, Hannah Sheahan$^*$, Christopher Summerfield, Lucy Campbell-Gillingham

\textbf{The model serving infrastructure} was designed and built by: Hannah Sheahan, Martin Chadwick, Michiel Bakker, Amelia Glaese, John Aslanides

\textbf{The debate question dataset} was created by: Michiel Bakker, Christopher Summerfield

\textbf{Data analysis and visualisation}: MH Tessler, Michiel Bakker, Martin Chadwick, Hannah Sheahan, Christopher Summerfield

\textbf{Writing}: Michiel Bakker$^*$, MH Tessler$^*$, Martin Chadwick, Christopher Summerfield, Hannah Sheahan

\textbf{Operations}: Lucy Campbell-Gillingham

\textbf{Conceptualization}: Michiel Bakker, Martin Chadwick, Hannah Sheahan, Christopher Summerfield, Matthew Botvinick

$^*$Authors contributed equally
\bibliographystyle{plain}
\bibliography{citation}

\newpage

\appendix

\setcounter{table}{0}
\renewcommand{\thetable}{A\arabic{table}}
\renewcommand*{\theHtable}{\thetable}
\setcounter{figure}{0}                       %
\renewcommand\thefigure{A\arabic{figure}}   %

\section*{Appendix}

\begin{table}[t]
    \caption{Prompt templates for question generation, zero-shot prompting, few-shot prompting and reward modelling. For supervised fine-tuning (SFT), we use the same prompt template as for zero-shot prompting. Note that, for reward modelling, we are not generating text but rather feed the embeddings after the final token to an additional linear layer.}
    \label{table:prompt_templates}
    \centering
    \footnotesize
    \begin{tabular}{@{}p{0.77in}p{\dimexpr \linewidth-2\tabcolsep-0.77in}@{}}
    \toprule
    Use case & Prompt template \\
    \midrule \midrule
    Question \newline generation & \texttt{An intelligent computer system is constructed. It is friendly and safe. The system generates debate questions that can be used to have interesting discussions between people. The questions the system came up with are: \newline \newline Question: [sampled seed question 1] \newline Question: [sampled seed question 2] \newline …  \newline Question: [sampled seed question 10] \newline [RESPONSE]}\\ \midrule 
    Zero-shot prompting and supervised fine-tuning (SFT) & \texttt{A citizen's jury was tasked with coming up with consensus opinions on a range of different questions. Below we present one such question along with the opinions of each individual citizen followed by their consensus statement.\newline \newline Question: [debate question] \newline Opinion: [human opinion 1] \newline ...  \newline Opinion: [human opinion N] \newline \newline After a good debate, the citizen's jury came to the following consensus view: [RESPONSE]} \\ \midrule
    Few-shot \newline prompting & \texttt{A citizen's jury was tasked with coming up with consensus opinions on a range of different questions. Below we list these questions along with the opinions of each individual citizen followed by their consensus statement. \newline \newline Question: [example debate question 1] \newline Opinion: [example human opinion 1.1] \newline ...  \newline Opinion: [example human opinion 1.N] \newline Consensus: [example consensus 1] \newline \newline ... \newline \newline Question: [example debate question M] \newline Opinion: [example human opinion M.1] \newline ...  \newline Opinion: [example human opinion M.N] \newline Consensus: [example consensus M] \newline \newline Question: [debate question] \newline Opinion: [human opinion 1] \newline ...  \newline Opinion: [human opinion N] \newline Consensus: [RESPONSE]}  \\ \midrule
    Reward \newline modelling & \texttt{Question: [debate question] \newline Opinion: [human opinion] \newline Consensus: [candidate consensus]}  \\ \bottomrule
    \end{tabular}
\end{table}

\section{Generating debate questions}

\begin{table}[t]
    \caption{Cluster topic and example questions from a subset of clusters in the within-distribution question set.}
    \label{table:example_questions_by_cluster}
    \centering
    \footnotesize
    \begin{tabular}{@{}p{0.8in}p{\dimexpr \linewidth-2\tabcolsep-0.8in}@{}}
    \toprule
    Cluster & Example questions\\
    \midrule \midrule
    9: Smoking & \texttt{Should the sale of cigarettes be banned? \newline 
    Should we ban the sale of tobacco products altogether? \newline
    Should we ban the selling of e-cigarettes?
    }\\ \midrule 
    15: Prisons & \texttt{Should prison be less comfortable? \newline
    Should we replace prisons with rehabilitation programs? \newline
    Should we ban private prisons?
    }\\ \midrule
    44: Plastics & \texttt{Should we ban the sale of single-use plastic items? \newline
    Should we ban the use of non-biodegradable plastic bags? \newline
    Should we ban plastic-based microbeads in cosmetic products?} \\ \midrule
    57: Vaccination & \texttt{Should we require mandatory vaccinations? \newline 
    Should all children be vaccinated against preventable diseases? \newline
    Should the government be responsible for public health? 
    } \\
    \bottomrule
    \end{tabular}
\end{table}

\subsection{Generating questions from seed questions}
We need access to a large and diverse dataset of debate questions to train and test our models. To generate these questions, we use the pre-trained \emph{Chinchilla} model using few-shot prompting and 152 hand-written seed questions \cite{hoffmann2022training}. We go through the following process to generate each question:
\begin{enumerate}
    \item We randomly sample 10 seed questions from the corpus of 152 hand-written seed questions. See Table~\ref{tab:seed_questions} for 10 examples of seed questions.
    \item We prompt the model with the 10 sampled seed questions, each delimited by a new line token '\textbackslash\textit{n}'. See the prompt template in Table~\ref{table:prompt_templates}. We use nucleus sampling with a cut-off probability of $0.8$ and a temperature of $1$ \cite{holtzman2019curious}.
    \item We then take the generated output, remove the \texttt{Question: }prefix and truncate everything after the first new line token.
    \item If the generated question is longer than 15 characters, contains the word ``should'' and is not in our dataset yet, we add this question to our dataset of questions.
    \item We repeat this process until we have 3500 questions in our dataset.
\end{enumerate}
Note that the seed questions are not directly used in the final question set and that the newly generated questions are not used as additional seed questions.

\begin{table}[t]
    \caption{Ten example seed questions selected from the 152 seed questions.}
    \centering
    \footnotesize
    \begin{tabular}{l}
    \toprule
         Example seed questions \\
         \midrule
         \texttt{Should we adopt blasphemy laws?} \\
         \texttt{Should we abandon the idea of HS2?} \\
         \texttt{Should we prevent MPs from having second jobs?} \\
         \texttt{Should short haul flights be banned within the UK?} \\
         \texttt{Should we cut the subsidy to the BBC?} \\
         \texttt{Should trans fat usage in food be banned?} \\
         \texttt{Should the British monarch not be allowed to issue a royal pardon?} \\
         \texttt{Should health care be free to everyone at the point of care?} \\
         \texttt{Should we support water privatization?} \\
         \texttt{Should we subsidize the cost of home insulation?} \\
   \bottomrule
    \end{tabular}

    \label{tab:seed_questions}
\end{table}

\subsection{Curation process}
To ensure that questions are safe and effective and can be used to elicit diverse opinions, we check each of the 3500 generated questions manually and exclude questions for any of the following seven reasons:
\begin{enumerate}
    \item The question is nonsensical or logically incongruent.
    \item The question requires the reader to understand technical jargon unlikely to be in the public vocabulary.
    \item We excluded questions that we felt might be likely to provoke extremist views or discriminatory attitudes, in particular where the discussion would be at risk of differentially impacting protected groups, including age, gender reassignment, being married or in a civil partnership, being pregnant or on maternity leave, disability, race including colour, nationality, ethnic or national origin, religion or belief, sex, and sexual orientation.
    \item The question proposes violence against someone or a group of individuals.
    \item The question refers to decisions that other countries or populations should make and hence is not targeted at our target demographic (UK citizens). 
    \item The question will not lead to a diverse discussion by virtue of being either too specific or too broad to have reasonable disagreement.
    \item The question contains one or more false presuppositions.
\end{enumerate}
After curation, there were 2922 questions remaining.

\subsection{Topic clustering and out-of-distribution question set}
\begin{figure}
    \centering
    \includegraphics[width=\textwidth,trim={0 7.4cm 0 0},clip]{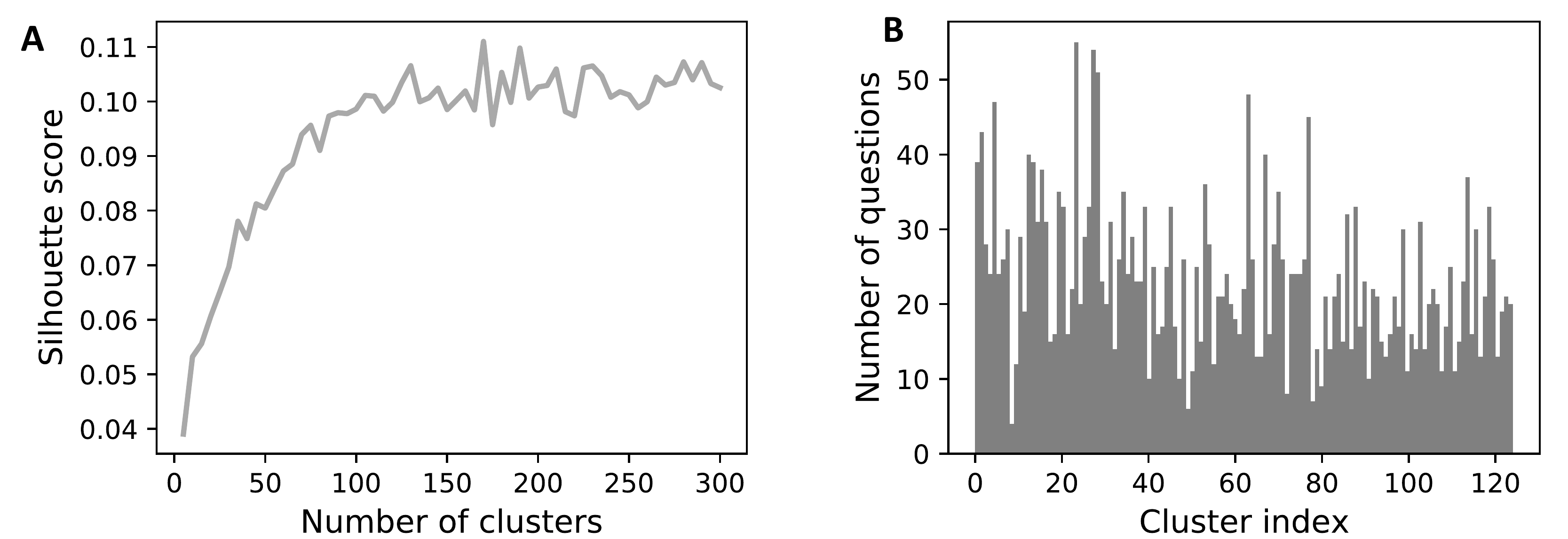}
    \caption{K-means clustering of the debate question embeddings. A: The Silhouette score as a function of the number of clusters. B: The distribution of questions per cluster (total of 2922 questions).}
    \label{fig:clustering}
\end{figure}
We split the remaining 2922 questions by topic using k-means clustering, and allocate some clusters to a special out-of-distribution test set. In this way we can test how our model generalises to topics that are unseen at training time. To identify the topic of each question, we use a Universal Sentence Encoder \cite{cer2018universal} with k-means clustering and follow the steps below:
\begin{enumerate}
    \item First, remove phrases that occur in many questions and are irrelevant for clustering into topics. This includes the following strings: ``should we'', ``should the government'', ``government'', ``the uk'', ``the united kingdom'', ``britain'', and ``great britain''.
    \item We encode each question using a Universal Sentence Encoder (USE) via Tensforlow Hub (using the following module: \texttt{https://tfhub.dev/google/universal-sentence-encoder/4}). This maps each question to a 512-dimensional embedding.
    \item Using the standard \texttt{sklearn} k-means implementation, we cluster the USE embeddings while sweeping the number of total clusters from 5 to 300 in increments of 5. For each number of clusters, we randomly reinitialize 100 different times, and keep the clustering that that produces the highest Silhouette score. We find that the Silhouette score converges at 125 clusters (Figure~\ref{fig:clustering}A). We use these 125 clusters for choosing train and test question sets and show the distribution of questions over these clusters in  Figure~\ref{fig:clustering}B.
    \item We then randomly allocate some clusters one-by-one to an out-of-distribution question set, and continue adding clusters until we have at least 300 examples in this set. To ensure similarity of topic-spread between the within-distribution and out-of-distribution question sets, we create 1000 random splits between the within-distribution and out-of-distribution set and measure the mean Euclidean distance between clusters in both sets. We then choose the within-distribution and out-of-distribution sets as those that have the most similar mean cluster distance. The final out-of-distribution test set has 302 questions across 14 different topics. 
\end{enumerate}
We refer to Table~\ref{table:example_questions_by_cluster} for example questions from each a subset of clusters. Finally, the within-distribution question set is split into a training set and a within-distribution test set, which contains questions which are different from the training set but belong to the same clusters. 
\section{Human data collection}
\subsection{Experimental protocol}
\begin{figure}
    \centering
    \includegraphics[width=\textwidth]{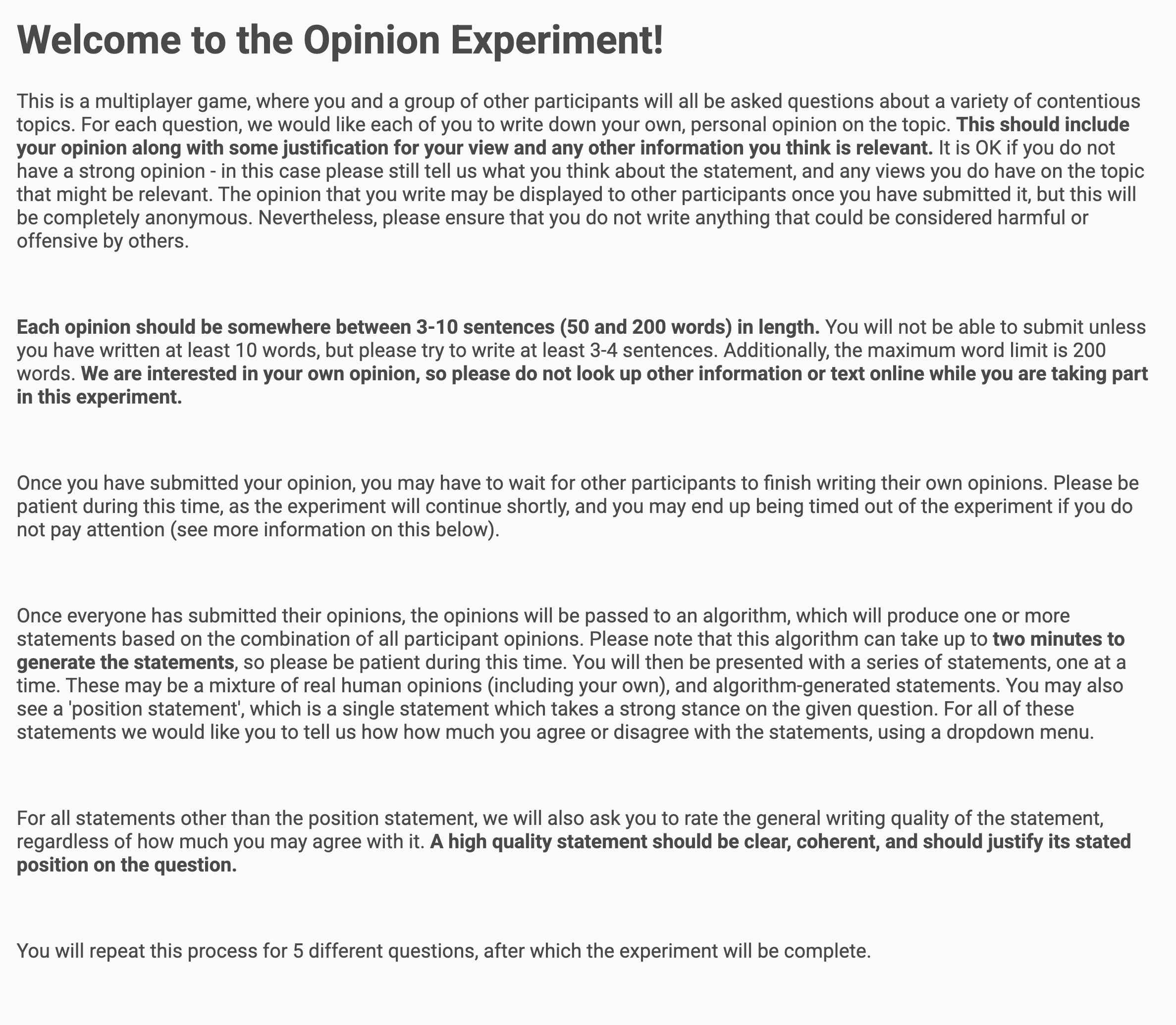}
    \caption{Main task instructions provided to all participants. Additional information (not shown) was provided regarding the task time limits and the button for reporting any offensive or inappropriate content.}
    \label{fig:instructions}
\end{figure}

All participants were recruited via an online crowd-sourcing platform, and took part in our experiments via a web interface. Human data was collected for both training and evaluation using the same basic experimental protocol and custom interface, which we describe here in detail. Each participant first read the task instructions (see Figure~\ref{fig:instructions}), and completed a short comprehension test. The comprehension check was designed to test the participants' knowledge and understanding of key aspects of the experiment. Each question was multiple choice, and after an answer was selected we provided immediate feedback to consolidate or correct the participant's understanding. We did not exclude any participants at this point. Upon completion of the comprehension test, participants were redirected to the experiment lobby, where they were asked to wait until enough other players had joined to form one experimental group (the size of which was always either four or five, depending on the particular experiment - details are in the main text). Once all players had joined, the group started the main experiment. In practice, data was collected in batches of around 20 groups (100 participants) in parallel.

Each experiment consisted of four rounds, each of which was based on a different political question. Within each round, there were three consecutive phases: opinion writing, candidate generation, and candidate rating. First, participants would view the question of interest, and would be provided with a text box in which they were asked to write their own opinion in response to the question. Participants were explicitly asked to write their own opinion on the topic, and not to disengage from the experiment in order to read any external sources. Pasting was disabled in the text box to further discourage the use of external opinions. Opinions had to be between 10 and 200 words in length and participants were encouraged to write between 3 and 10 sentences. As this was a multiplayer game, we applied a time limit of 5 minutes to the opinion writing to ensure that participants were not forced to wait too long for others to finish. The time remaining was always displayed at the top of the page during this phase. If a participant failed to complete within this time limit, the game moved on to the next phase, and their opinion was not included in the candidate generation process. Once all participants in a group had submitted their written opinions, these were passed to one or more models, and used to generate a set of consensus candidates. During this phase participants' viewed a waiting page which could last between 20 to 120 seconds depending on the particular experiment.

\begin{figure}
    \centering
    \includegraphics[width=\textwidth]{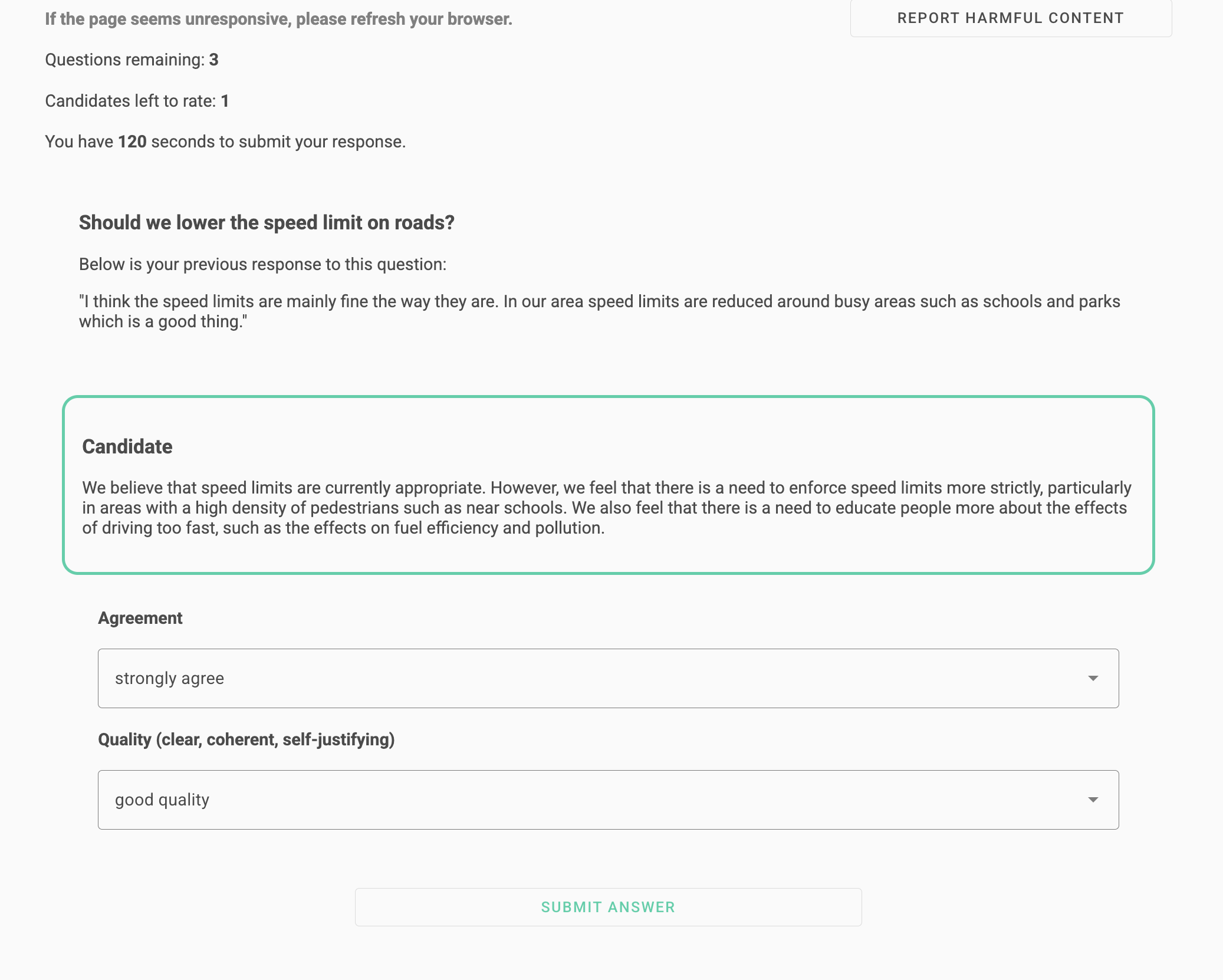}
    \caption{User interface displayed to participants when providing ratings for candidate consensus statements.}
    \label{fig:interface}
\end{figure}

Once the candidates had been generated, participants progressed to the final phase - candidate rating. The user interface for this phase is displayed in Figure~\ref{fig:interface}. Both the question and the participant's own written opinion were displayed as a reminder, and the consensus candidate was displayed beneath. Two drop-down menus allowed participants to provide seven-point Likert scores for both agreement and quality (full labels are provided in Table~\ref{tab:likert_table}). Prior to rating the model or human-generated candidates, participants were asked to provide an agreement (but not quality) rating to a 'position statement', which was simply the question itself restated as a declarative statement (e.g. The position statement for the question 'Should we lower the speed limit on roads?' was 'We should lower the speed limit on roads.'). This initial rating provided us with a general indication of the range of opinions to the question, independent of the candidate generation process. We used these statements primarily for determining the divisiveness of each question (see main text and section below).

Following the position statement, all consensus candidates were presented sequentially. All were presented once in random order, then again in a new random order. This repetition of each statement allowed us to measure the intra-rater reliability, which we used for data quality filtering (see Section~\ref{sec:irr}). Our intention here was to filter out ratings from participants that often varied hugely between their first and second ratings, as we consider this to be a proxy for the participant paying attention to the task (rather than responding randomly). We expected participants' two responses to be roughly the same, with some tolerance for noise, which informed our usage of a relatively lenient intra-rater reliability threshold. As with the opinion writing phase, participants were given a time limit to complete all of the ratings during the ratings phase. The precise limit varied between experiments, as different experiments presented different numbers of candidates. In all cases, the time limit was based on allowing approximately 60s per candidate. Time remaining was displayed at the top of the page throughout this phase, along with additional information regarding the number of questions remaining, and candidates left to rate. If a participant ran out of time more than once, we removed them from the experiment and they were still compensated at the full rate.

Additionally, we provided a button labelled 'Report Harmful Content', which participants were instructed they should press if they were exposed to any content that they deemed to be offensive or inappropriate. In the event that this button was pressed, participants would be taken out of the experiment and would be asked to (optionally) provide us with further information about the harmful content via a text box. Whenever this occurred, participants were compensated at the full rate for the experiment. However, we note that no participants reported any harmful content (full details in section below).

\begin{table}[t]
    \centering
    \footnotesize
    \caption{Qualitative labels for the two Likert scales.}
    \begin{tabular}{||c c c||} 
    \hline
    Score & Agreement & Quality\\  
    \hline\hline
    7 & \texttt{Strongly Agree} & \texttt{Excellent Quality} \\ 
    \hline
    6 & \texttt{Agree} & \texttt{Good Quality} \\
    \hline
    5 & \texttt{Somewhat Agree} & \texttt{Somewhat Good Quality} \\
    \hline
    4 & \texttt{Neutral} & \texttt{Neutral} \\
    \hline
    3 & \texttt{Somewhat Disagree} & \texttt{Somewhat Poor Quality}  \\ 
    \hline
    2 & \texttt{Disagree} & \texttt{Poor Quality}  \\ 
    \hline
    1 & \texttt{Strongly Disagree} & \texttt{Terrible Quality}  \\ 
    \hline
    \end{tabular}
    \label{tab:likert_table}
\end{table}

\subsection{Quality control processes}
\subsubsection{Intra-rater reliability}\label{sec:irr}

\begin{figure}
    \centering
    \includegraphics[width=\textwidth]{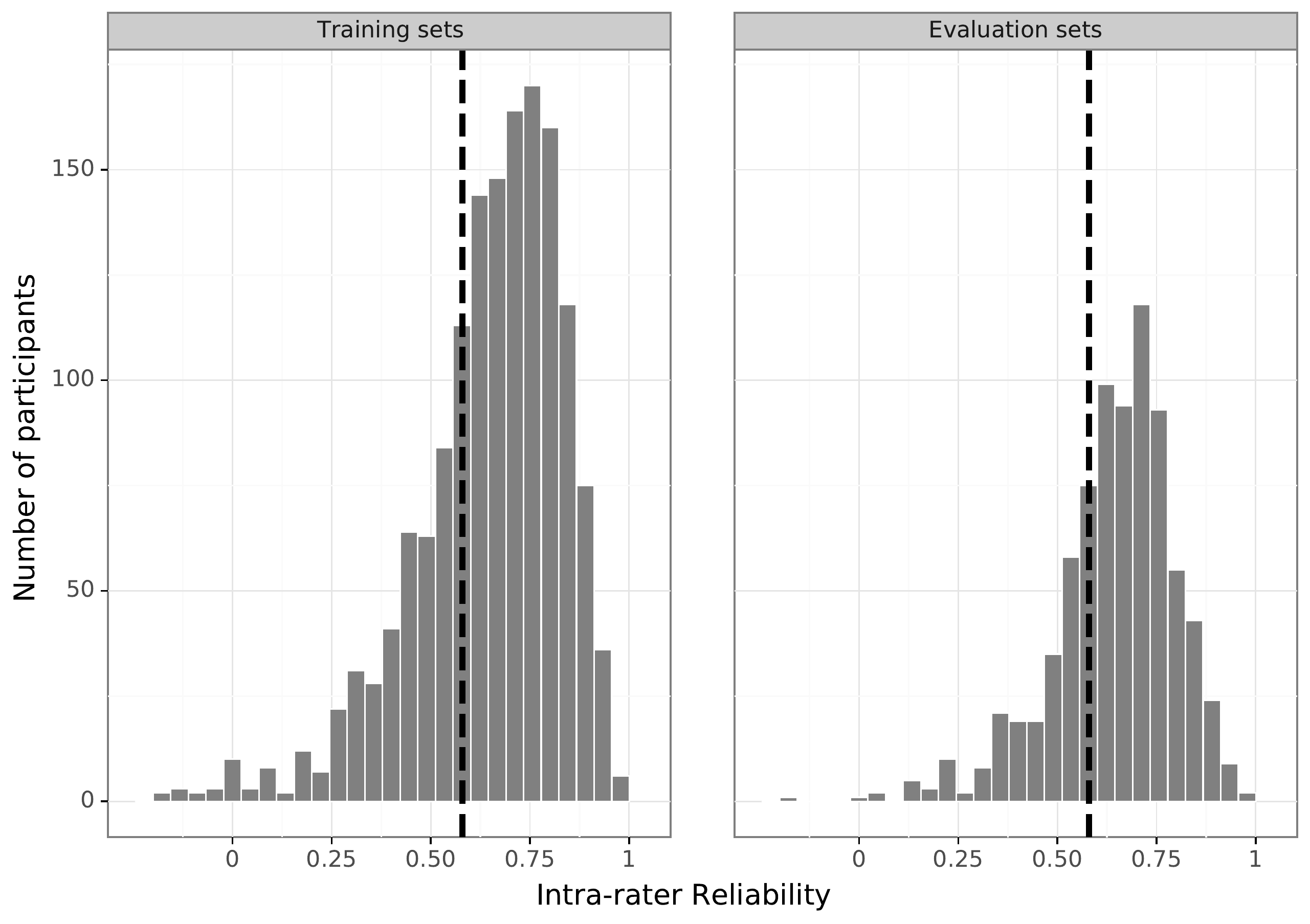}
    \caption{Intra-rater reliability using linearly-weighted Cohen's Kappa of Training data and Evaluation data. Vertical line shows our threshold for exclusion at $\kappa = 0.6$.}
    \label{fig:irr}
\end{figure}

Since participants provided two ratings (under each quality and agreement metric) for all model candidate consensus statements (and, human opinions, during the Human Opinion eval), we can compute an intra-rater reliability score for each participant. Intra-rater reliability provides a quantitative measure for how consistent participants' responses were across the two presentations of the same model statement (or human opinion). To measure intra-rater reliability, we computed Cohen's Kappa coefficient $\kappa$ using linear weights (i.e., linearly-weighted Cohen's Kappa), which allowed us to take into account the fact that the ratings provided are along an ordered scale. In early batches of training data, we observed a bimodal distribution for $\kappa$, with a threshold of approximately 0.6 separating the ``reliable participants'' from the ``unreliable participants'' (0.6 is  also commonly considered in the psychometric literature to be the threshold between ``moderate'' agreement and ``substantial'' agreement) \cite{mchugh2012interrater}. Therefore, we used 0.6 as a threshold for filtering unreliable participants. Using this threshold, we retained 1028/1524 (67\%) in our training data and  1164/1687 participants (69\%) in the evaluation data. The empirical distributions over $\kappa$ for the full training and evaluation data sets are shown in Figure \ref{fig:irr}.

\subsubsection{Harm report}
As described above, our user interface allowed participants to report any instances of offensive or inappropriate content being generated by the model, or indeed by other participants in conditions where participants viewed the opinions of others. An analysis of the data reveals that this functionality was used a total of nine times across all of our training and evaluation data collection, and in all cases this was explicitly reported to be a mistake by the participant. This indicates that over 746 separate groups of participants, our model did not generate content that was deemed sufficiently inappropriate or offensive to be flagged by the participants.

\section{Model training and evaluation}
We use two different language models, a generative model for the consensus candidates and a classifier to predict the reward. The models are fine-tuned after each of the two consecutive rounds of data collection. The first round of data collection corresponds to participants that rate consensus candidates generated with zero-shot and few-shot prompting, while the second round corresponds to candidates generated using the models that were trained using the first round of data. For fine-tuning, we always start from a pre-trained \emph{Chinchilla} model both for the consensus generating model and for the reward model. Moreover, after the second round of data collection we fine-tune our models using a mix of data from the first and second round, in line with prior work (e.g. \cite{ziegler2019fine}). Given the high cost of data collection and model training, as well as the relatively small increase in model performance between the first and second round of training, we stop data collection after two rounds.

\subsection{Consensus generating model}
We use three different consensus generating models, a zero-shot and few-shot prompted \emph{Chinchilla} model, and a fine-tuned (SFT) \emph{Chinchilla} model. At inference time, we sample consensus candidates using nucleus sampling with a cut-off probability of $0.8$ and a temperature of $1$ \cite{holtzman2019curious}. When we use the reranking models (e.g. SFT-Utilitarian), we sample $16$ candidates and choose the sample that maximises the expected welfare. Note that the sampling happens live while participants wait for the candidate rating phase to begin. To increase candidate diversity and increase order-invariance, we use a different random order of opinions in the prompt each time we sample.

\subsubsection{Zero-shot and few-shot prompting}
We use two approaches for prompting the base LLM model (\emph{Chinchilla}). We refer to these two prompt methods as ``zero-shot'' and ``few-shot'' prompting. With zero-shot prompting, we simply provide a few sentences designed to elicit a prior based on the reports of citizens' juries, which we considered to be a good template for generating consensus statements. Following this, the prompt always included the specific question along with the human-provided opinions, but with no previous examples of what a consensus candidate statement should look like (see the zero-shot template in Table~\ref{table:prompt_templates}). For few-shot prompting, rather than create hand-crafted (and potentially unrepresentative) examples ourselves, we elected to collect some initial training data using the zero-shot approach. Once these initial batches were collected, we then created a generative few-shot prompting approach using real opinions and consensus candidates collected using the zero-shot prompting approach. By using real data for the examples in the prompt, the examples should be more representative of our participant population, while also allowing us to create sample diversity through our generative process. 

The few-shot prompt was similar to the zero-shot one, but with the addition of three examples (see the few-shot template in Table~\ref{table:prompt_templates}). Each example consisted of a question, some opinions, and a model-generated consensus candidate statement, all of which were drawn from data collected using the zero-shot prompted model. Every prompt included one example with three opinions, one with four, and one with five, in order to encourage diversity and generalization over different numbers of input opinions. Each candidate generated under this method was based on a unique prompt generated using the following method, applied separately to the 3, 4, and 5 opinion candidates:
\begin{itemize}
    \item Filter to include all examples where the candidates had a mean quality rating higher than ``Neutral’’.
    \item Rank all consensus candidates under the specified social welfare function.
    \item Sample a candidate from the top 10 ranked candidates.
    \item Construct the example using the selected candidate along with the associated question and opinions.
\end{itemize}

This process provided a combination of prompt diversity along with a degree of example optimization towards the specified social welfare function in each case.

\subsubsection{Supervised fine-tuning}
To further improve the consensus generating LLM over zero-shot and few-shot prompting, we fine-tune the pre-trained \emph{Chinchilla} model using supervised finetuning, in line with previous works \cite{menick2022teaching, thoppilan2022lamda, ziegler2019fine}. For supervised fine-tuning we use the same prompt as for zero-shot prompting, see Table~\ref{table:prompt_templates}. We also tried a simpler prompt without the first \texttt{A citizen’s jury was tasked...} sentence but find that adding the sentence makes fine-tuning more data-efficient and more reliable.
 
\paragraph{Training data}\label{sec:sft_training_data}
While some previous works rely on human demonstrations for supervised fine-tuning (SFT), we find that many of the candidates that were generated using zero-shot and few-shot prompting are of sufficient quality to be used for further fine-tuning. Hence, we use highly rated consensus candidates from previous rounds (generated via prompting in the first round and generated with the first iteration SFT model in the second round) as target for fine-tuning using the following steps:
\begin{enumerate}
    \item We first split the training data by question and allocate $80\%$ of the questions to a training set and $20\%$ to a validation set which we use for hyperparameter optimisation and early-stopping.
    \item For the training data we then filter out the high-quality candidates as rated by our participants. These targets correspond to consensus candidates that have a mean rating higher than $5$ (corresponding to ``Somewhat Good Quality’’), averaged over reliable participants (see Section~\ref{sec:irr}). For the validation set, we only use candidates that have a mean rating higher than $6$  (corresponding to ``Good Quality’’) to ensure that we select the highest-quality model. For training, we explored the quality-quantity trade-off by using different minimum mean rating thresholds ($\{4, 4.5, 5, 5.5, 6\}$) to build the training set but found that a threshold of $5$ leads to the best performance on the validation set. 
    \item We further ``clean’’ each candidate by 1) truncating any text after the first new line character 2) removing text that is repeated from the prompt (e.g. the model sometimes starts a consensus with \texttt{Opinion:}), 3) removing any leading or trailing quotation marks and whitespace.
    \item Finally, to encourage order-invariance with respect to the opinions and increase diversity during the training process, we randomise the order of the opinions in the prompt at start of every new training epoch.
\end{enumerate}
This process yields 1525 candidates for training in the first round of fine-tuning and 2355 candidates in the second round of fine-tuning. Note that in the second round, we use the examples from both rounds and thus use 3880 candidates in total.

\paragraph{Training and evaluation}
During training, we evaluate our model performance using perplexity on the evaluation set, a measure corresponding to the inverse probability of predicting the test set using the current model. We stop training when the perplexity on the evaluation set no longer decreases. In both training rounds, we use a learning rate of $2.5 \cdot 10^{-7}$, a batch size of 256, and the AdamW optimiser \cite{loshchilov2017decoupled}. To fit the 70 billion parameter model on TPU memory, we shard the model over 64 TPU v3 cores. We train all layers except the embedding layer and train in low precision using bfloat16 with stochastic rounding \cite{menick2022teaching,hoffmann2022training}.

In the first round, we train for 85 steps (14.3 epochs) and in the second round for 145 steps (9.6 epochs) before the model starts to overfit. Note that we train for more epochs than some other recent work that fine-tunes LLMs \cite{menick2022teaching}. This can be explained by the fact that the order of the opinions in the prompt is randomised between epochs and the model thus rarely sees the exact same example multiple times during training. This also leads to a lower optimal learning rate compared to most previous work. We tried both higher and lower learning rates but found that  $2.5 \cdot 10^{-7}$ yields the lowest evaluation perplexity.

In Figure~\ref{fig:training_figs}A, we compare the perplexity for the SFT models to zero-shot and few-shot baselines using the data that was collected using the within-distribution and out-of-distribution test questions. We observe a performance increase both by going from prompting to SFT but also between both the two rounds of SFT. 

\subsection{Reward model}
In addition to the consensus generating model, we train a reward model that takes in the question, a single opinion and a consensus candidate and predicts a scalar reward. The inputs are combined using the prompt template for the reward model shown in Table~\ref{table:prompt_templates}. At inference time, we use the scalar rewards for each opinion to select the consensus candidate that is expected to maximise group welfare.

\paragraph{Training data}
Following Christiano et al. \cite{christiano2017deep}, we train our reward model by predicting the preferred candidate using pairwise preferences. In our case, we extract those pairwise preferences from the agreement Likert ratings given by participants using the following process:
\begin{enumerate}
\item In line with SFT, we first split our data by question and allocate 80\% of the questions to a training set and 20\% to a validation set. 
\item We extract all the possible pairwise preferences after removing ratings from unreliable raters.
\item We filter out all the rating pairs for which the difference in ratings is 1 point or less on the 7-point Likert scale. We also tried filtering out only strict ties (so allowing for example an (``Agree’’,``Strongly Agree’’) pair) but find empirically that being stricter on which pairs we include leads to faster training and better evaluation performance, even when we use an evaluation set for which we filter out only strict ties.
\item Finally, before putting the consensus candidate in the prompt template that is fed to the reward model, we ``clean’’ the candidate in a similar way as for SFT (see Section~\ref{sec:sft_training_data}).
\end{enumerate}

This process yields $6639$ rating pairs for reward model training in the first round and $3795$ rating pairs in the second round. Note that, in line with SFT, we use the examples from both rounds ($10434$ pairs) for training the second iteration of the reward model.

\paragraph{Training and evaluation}
At training time, we warm-start the reward model using the pre-trained \emph{Chinchilla} model with an additional final linear layer to predict the reward. Some previous work has warm-started the RM with an SFT model. However, in our case the prompt templates are too different for this to benefit performance. Because we are predicting which candidate consensus is preferred out of a pair, we train and evaluate our models using a standard cross-entropy loss. We stop training when the evaluation loss no longer decreases. In both training rounds, we use a maximum learning rate of $2 \cdot 10^{-6}$ with a linear warmup and cosine anneal, a batch size of 8, and the Adam optimiser. To fit the 70 billion parameter model on TPUs, we shard the model over 32 TPU v3 cores. For the RM we only train the last 25 layers (including the extra linear layer) and, in line with SFT, we train in low precision using bfloat16 with stochastic rounding. To stabilise training during the beginning of training, we first freeze the pre-trained language model layers for 0.1 epochs and only train the extra linear layer during those steps. 

In the first round, we train for 440 steps (0.5 epochs) and in the second round for 1113 steps (0.8 epochs) before the model starts overfitting. Note that, in contrast to SFT, we now train for less than one epoch which can be explained by the fact that we have extracted all possible combinations of pairwise ratings leading to many correlated pairs that share candidates. Hence, even though the model has not seen every unique rating pair during training, it has seen the vast majority of candidates multiple times.

\begin{figure}
    \centering
    \includegraphics[width=\textwidth]{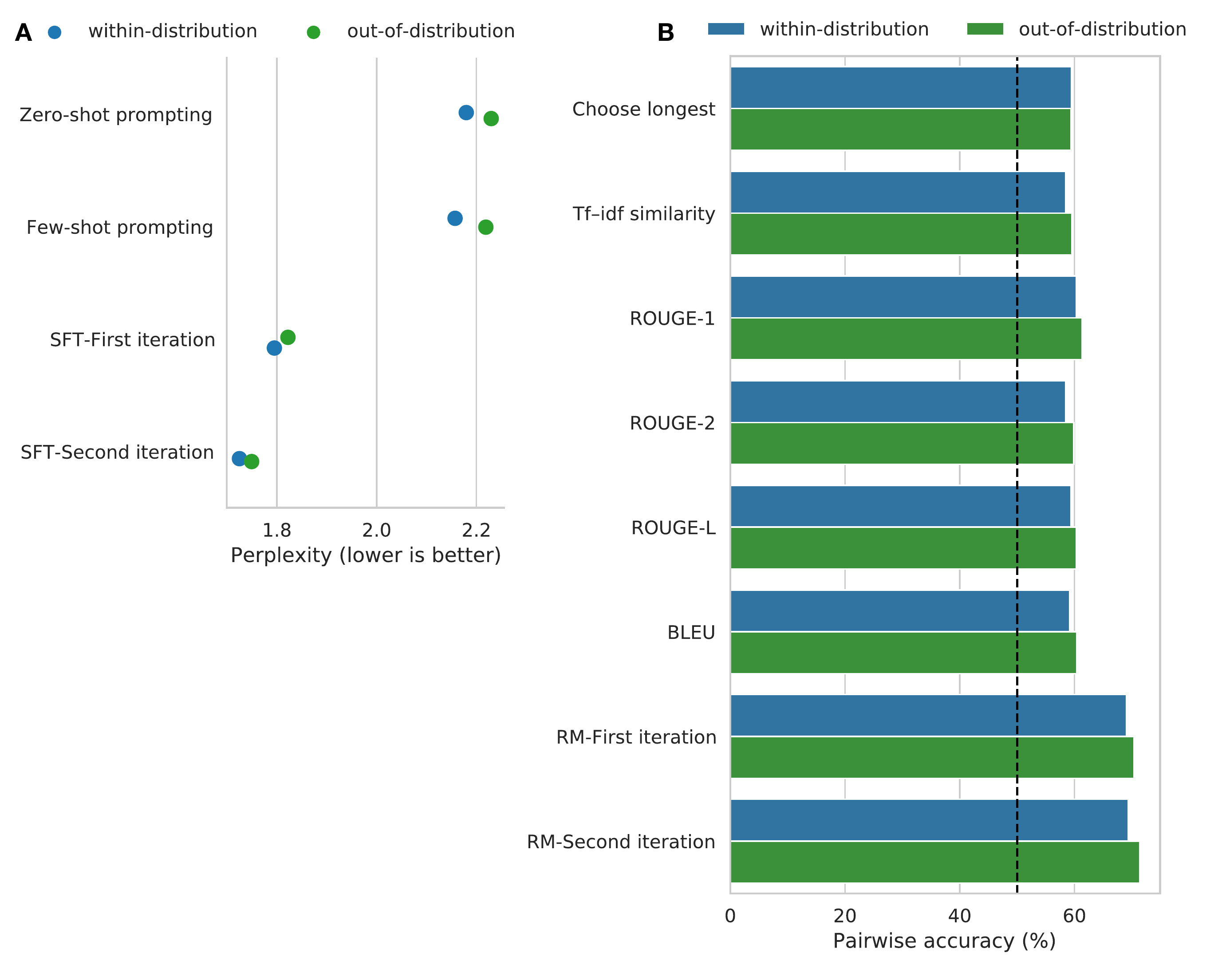}
    \caption{Model performance evaluated on the data that was collected for comparing SFT-Utilitarian to baseline models, using the within-distribution and out-of-distribution questions (see Sections~4.1 and~4.2  in the main paper). Given the size of the models, we train using only a single random seed and do not show error bars.  A: Consensus generating model perplexity for the first and second SFT model compared to zero-shot and few-shot prompted. For this evaluation, only candidates with a mean rating higher than $6$ were included in the evaluation set. B: Reward model performance in terms of pairwise accuracy (how well can the model predict which consensus candidate was preferred by a participant given their opinion and the question) compared to six baselines (see main text). For this evaluation, we use all the pairwise comparisons except when there is a strict rating tie.}
    \label{fig:training_figs}
\end{figure}

To benchmark the performance of the final model, we compare in Figure~\ref{fig:training_figs}b the pairwise accuracy (how often it selects the preferred candidate) for the final model, the reward model after the first training iteration, and six naive baselines:
\begin{itemize}
\item \textbf{Tf-idf with cosine similarity.} We train a Term Frequency Inverse Document Frequency (tf-idf) vectorizer on all the training data from both rounds and use this to vectorise the opinions and candidates. We then compute the cosine similarity between the opinion and each of the candidates in a pair and select the candidate that maximises this similarity as the preferred candidate. We use the standard \texttt{sklearn} tf-idf implementation with English stopwords removed. To optimize the model, we sweep which n-grams are extracted (from only uni-grams to uni-grams, bi-grams and tri-grams), and we sweep the upper and lower thresholds for which n-grams to include based on how often they occur.
\item \textbf{Choosing the longest.} We always select the candidate with the most characters as the preferred candidate, independent of the opinion or question.
\item  \textbf{ROUGE.} We always select the candidate with the highest ROUGE-1/ROUGE-2/ROUGE-L score \cite{ranzato2016sequence}. ROUGE (Recall-Oriented Understudy for Gisting Evaluation) is a set of metrics for evaluating summaries and machine translation. We compute ROUGE by looking at the overlap of the candidates and the opinion. ROUGE-1 corresponds to unigram overlap, ROUGE-2 to bigram overlap, and ROUGE-L to the longest common subsequence.
\item  \textbf{BLEU.} We always select the candidate with the highest BLEU score \cite{ranzato2016sequence}. BLEU (bilingual evaluation understudy) is a metric for evaluating machine translation. BLEU also computes the n-gram overlap but, in contrast to ROUGE, takes several size of n-grams into account simultaneously. Additionally, it introduces a brevity penalty.
\end{itemize} 

For benchmarking, we use the data that was collected using the within-distribution and out-of-distribution test questions. In Figure~\ref{fig:training_figs}b, we observe that both reward models outperform the naive baselines. Moreover, we observe a slight ($\sim1\%$) increase in accuracy between the first and second round of reward model training.

\section{Additional human data evaluations}

\subsection{Model size ablation experiment}

\begin{figure}
    \centering
    \includegraphics[width=\textwidth]{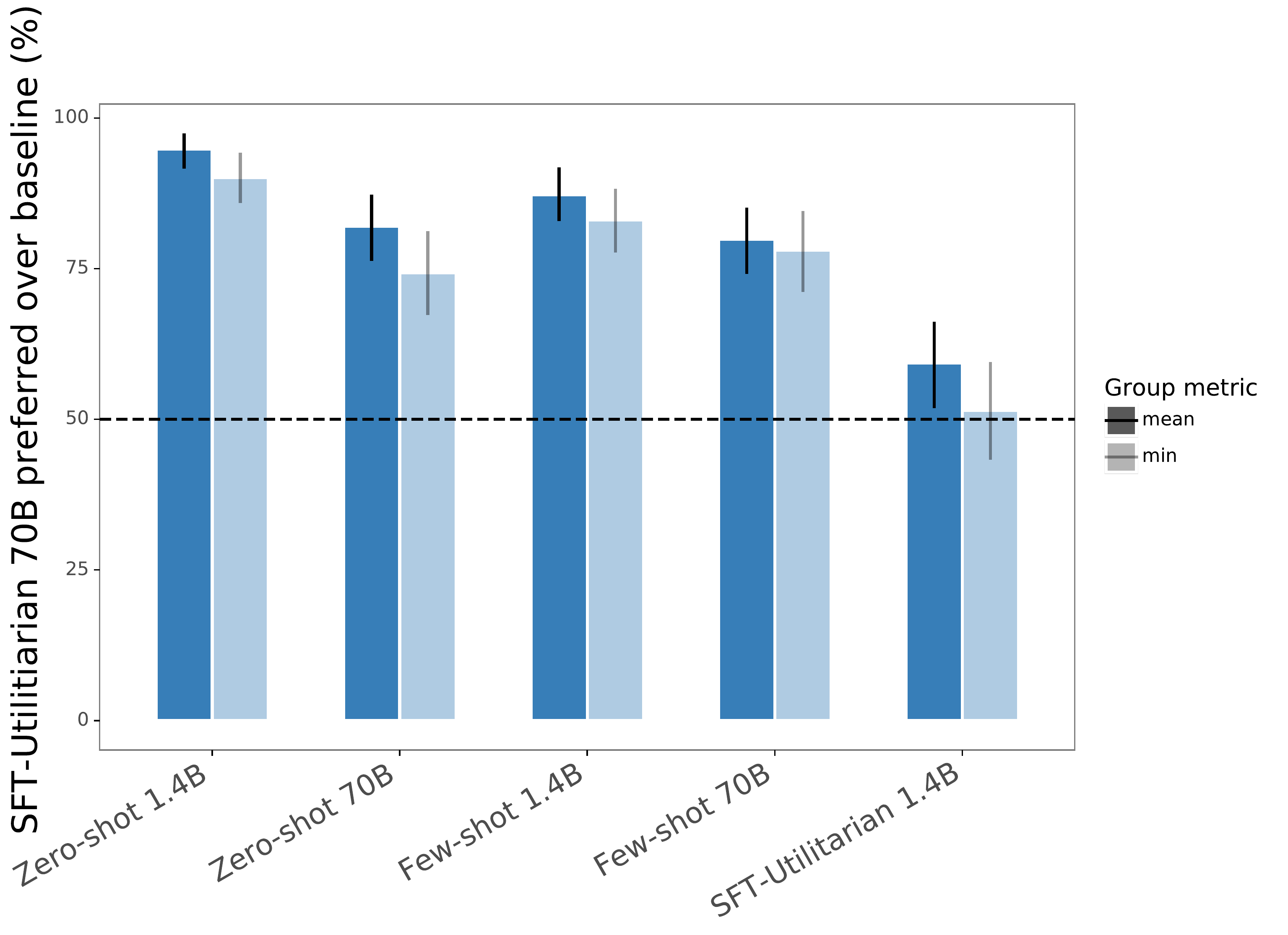}
    \caption{Preference win-rates for SFT-Utilitarian 70B vs. a set of 1.4B model baselines for the within-distribution question set.}
    \label{fig:SizeAblation}
\end{figure}

\begin{figure}
    \centering
    \includegraphics[width=\textwidth]{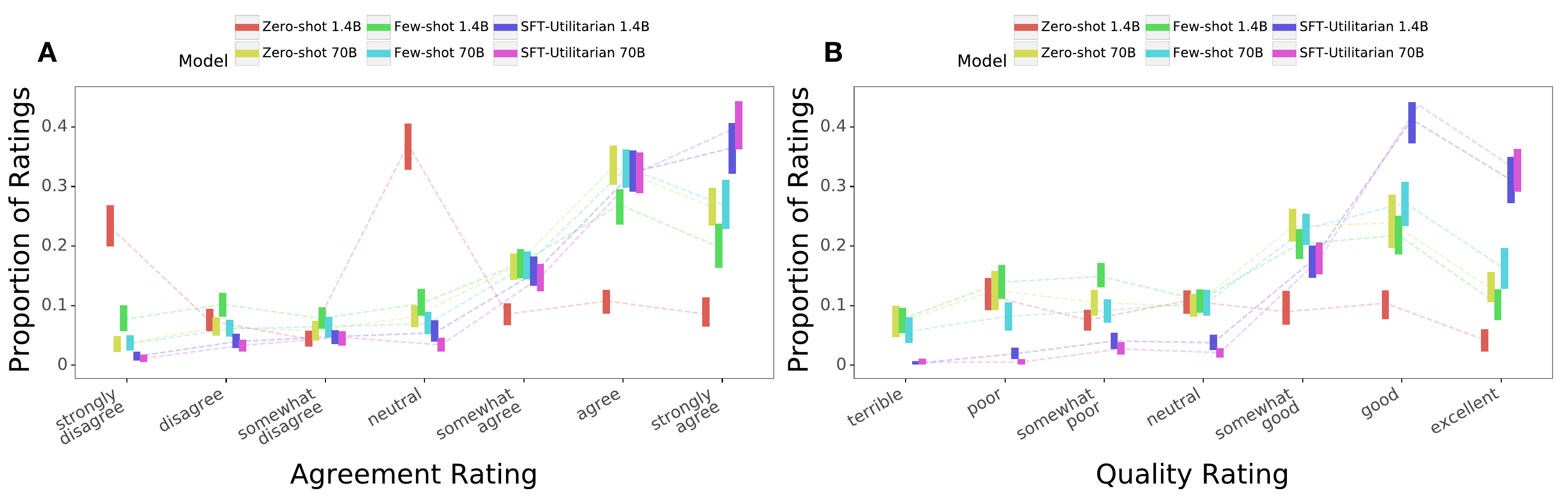}
    \caption{Distributions over Likert ratings for consensus statements generated by the SFT-Utilitarian and prompted baseline model-types, based on language models with 70B and 1.4B parameters. A: Agreement ratings. B: Quality ratings. Error-bars represent 95\% bootstrapped confidence intervals.}
    \label{fig:likerts-scale}
\end{figure}

In order to assess the impact of model size, we directly compared our main 70B parameter model against a smaller 1.4B model based on the same architecture and dataset as \textit{Chinchilla} \cite{hoffmann2022training}. We fine-tuned this model and trained a 1.4B reward model using the data we previously collected with the larger model. We conducted a human evaluation experiment ($n=224$), where we directly compared consensus candidates generated by 6 models: i) SFT-Utilitarian 70B, ii) Few-shot 70B, iii) Zero-shot 70B, iv) SFT-Utilitarian 1.4B, v) Few-shot 1.4B, and vi) Zero-shot 1.4B. As shown in Figure~\ref{fig:SizeAblation} SFT-Utilitarian 70B still significantly outperforms each baseline in win rate when comparing mean agreement. Interestingly, however, SFT-Utilitarian 1.4B performs only slightly worse (the 70B version wins over the 1.4B version 59.4\% [53.7\%, 64.7\%] of the time). This effect size is similar to SFT-Utilitarian 70B versus SFT-Base 70B, suggesting that both scale and reward modelling are independently beneficial for consensus generation.
Additionally, SFT-Utilitarian 1.4B outperforms both zero-shot and few-shot 70B (win rate of 75.9\% and 75.2\% respectively see Table~\ref{table:SizeAblationFull}). Notably, the smaller model was trained based on the data we previously generated using the 70B model. We speculate that the relative high-performance of the smaller fine-tuned model may be driven by the high-quality data generated by the larger model. Given the weaker prompting capabilities of the smaller model (see Table~\ref{table:SizeAblationFull}), it is likely that running the full iterative data collection pipeline from scratch using only the smaller model would have been more data intensive. This raises the interesting prospect of future, more efficient hybrid approaches, where initial data is generated using prompted large models, with fine-tuning applied to smaller models. Likert agreement rates for both small and large models are displayed in Figure~\ref{fig:likerts-scale}, while example consensus candidates are provided in Table~\ref{table:ScaleAblationExample}.

\begin{table}
    \caption{Pairwise win rates, comparing means, for models that vary in size (1.4B or 70B), and training pipeline (zero-shot, few-shot or finetuned). Both increasing model size and fine-tuning (SFT and reranking) independently improve performance. The SFT-Utilitarian 70B model is preferred over all others.}
    \begin{tabular}{@{}lll@{}}
    \toprule
    Model 1              & Model 2              & Win-rate 1 vs 2             \\ \midrule
    SFT-Utilitarian 70B  & SFT-Utilitarian 1.4B & 59.4\% {[}52.3\%, 66.7\%{]} \\
    SFT-Utilitarian 70B  & Few-shot 70B         & 79.9\% {[}73.9\%, 85.6\%{]} \\
    SFT-Utilitarian 70B  & Zero-shot 70B        & 82.0\% {[}76.6\%, 87.5\%{]} \\
    SFT-Utilitarian 70B  & Few-shot 1.4B        & 87.2\% {[}82.9\%, 91.5\%{]} \\
    SFT-Utilitarian 70B  & Zero-shot 1.4B       & 94.8\% {[}91.8\%, 97.5\%{]} \\
    Few-shot 70B         & Zero-shot 70B        & 51.4\% {[}44.9\%, 58.3\%{]} \\
    Few-shot 70B         & SFT-Utilitarian 1.4B & 24.8\% {[}18.5\%, 30.5\%{]} \\
    Few-shot 70B         & Few-shot 1.4B        & 65.5\% {[}59.6\%, 72.1\%{]} \\
    Few-shot 70B         & Zero-shot 1.4B       & 87.0\% {[}82.3\%, 91.0\%{]} \\
    Zero-shot 70B        & SFT-Utilitarian 1.4B & 24.1\% {[}18.7\%, 29.9\%{]} \\
    Zero-shot 70B        & Few-shot 1.4B        & 65.5\% {[}59.2\%, 72.4\%{]} \\
    Zero-shot 70B        & Zero-shot 1.4B       & 84.7\% {[}79.7\%, 89.3\%{]} \\
    SFT-Utilitarian 1.4B & Few-shot 1.4B        & 85.9\% {[}81.0\%, 90.4\%{]} \\
    SFT-Utilitarian 1.4B & Zero-shot 1.4B       & 92.3\% {[}88.8\%, 95.5\%{]} \\
    Few-shot 1.4B        & Zero-shot 1.4B       & 72.5\% {[}66.7\%, 78.7\%{]} \\ \bottomrule
    \end{tabular}
    \label{table:SizeAblationFull}
\end{table}

\subsection{Qualitative comparison of SFT-Utilitarian model to baselines}

See an example in Table~\ref{table:consensus_example}.

\begin{table}[t]
    \caption{An example of real participant opinions, model candidates and Likert scores in response to a question on smoking.}
    \label{table:consensus_example}
    \centering
    \footnotesize
    \begin{tabular}{@{}p{0.77in}p{\dimexpr \linewidth-2\tabcolsep-0.77in}@{}}
    \toprule
    \textbf{\normalsize Question} & Should smoking be banned in the home?\\
    \midrule \midrule
    \textbf{\normalsize Opinion 1} \newline \emph{participant} & No, this is completely impractical. It would be impossible to police. People should be encouraged not to smoke at all, and especially indoors in a place where other people live. It is especially important not to smoke near children. The message should be repeated frequently and people reminded at every opportunity. \\
    \midrule
    \textbf{\normalsize Opinion 2} \newline \emph{participant} & Banned in the home by who? The government? If so, the government has no place banning it in the home. In public spaces, yes, but someone's home is no place for the govt. to apply that rule. It's not their business and is crossing a certain personal boundary and freedom (also, practically, how would it be enforced in the home? cameras? neighbours reporting it?) \\
    \midrule
    \textbf{\normalsize Opinion 3} \newline \emph{participant} & Yes. Smoking is a bad habit and should be banned in general (both in the community and in people's houses). The reason for this is that smoking places a lot of pressure on our health services and smoking is a choice which people make. Therefore banning smoking will hopefully contribute towards an improvement around people's health. \\
    \midrule
    \textbf{\normalsize Opinion 4} \newline \emph{participant} & No. Whilst I think it should be banned around children, peoples home is their own private space and they should be allowed to make their own decisions. Banning smoking in people's private spaces is an affront to democracy. \\
    \midrule \midrule
    \textbf{\normalsize Consensus} \newline \emph{SFT-Utilitarian} & Smoking should be banned in public places. However, the home is a private space and people should be allowed to make their own decisions. Therefore, smoking should not be banned in the home.\\

      \emph{Participant 1} & Somewhat agree.\\
      \emph{Participant 2} & Strongly agree.\\
      \emph{Participant 3} & Somewhat disagree.\\
      \emph{Participant 4} & Strongly agree.\\
    \midrule
    \textbf{\normalsize Consensus}   \emph{SFT-Base} & Smoking should be banned in the home if there are children present. This is to protect children from the harms of passive smoking.\\      
      \emph{Participant 1} & Neutral.\\
      \emph{Participant 2} & Agree.\\
      \emph{Participant 3} & Disagree.\\
      \emph{Participant 4} & Neutral.\\
    \midrule
    \textbf{\normalsize Consensus} \newline \emph{Few-shot} & Yes, smoking should be banned in the home. This is for a number of reasons: 1) smoking is an expensive habit, 2) it can cause significant health problems (to both the smoker and to other people), 3) it is often difficult to stop, and 4) it is especially important to protect young children from smoke.\\

      \emph{Participant 1} & Disagree.\\
      \emph{Participant 2} & Somewhat disagree.\\
      \emph{Participant 3} & Agree.\\
      \emph{Participant 4} & Somewhat disagree.\\
    \midrule
    \textbf{\normalsize Consensus} \newline \emph{Zero-shot} & No. The citizen's jury agreed that smoking is a bad habit, but that banning smoking in peoples own homes is not an appropriate measure to take to reduce smoking. Whilst the jury agreed that smoking is a choice that people make, they also thought that banning smoking is not the right way to encourage people to change their habits.\\

      \emph{Participant 1} & Somewhat disagree.\\
      \emph{Participant 2} & Somewhat disagree.\\
      \emph{Participant 3} & Somewhat agree.\\
      \emph{Participant 4} & Somewhat disagree.\\
    \bottomrule
    \end{tabular}
\end{table}

\begin{table}[t]
    \caption{Consensus candidates generated with six different baseline models across two different model sizes (one with 70B and one with 1.4B parameters). The example was chosen randomly from the new human experiment after filtering for divisive questions (i.e., questions for which there was some disagreement in the group).}
    \label{table:ScaleAblationExample}
    \centering
    \footnotesize
    \begin{tabular}{@{}p{0.77in}p{\dimexpr \linewidth-2\tabcolsep-0.77in}@{}}
    \toprule
    \textbf{\normalsize Question} & Should the government subsidize house prices?\\
    \midrule \midrule
    \textbf{\normalsize Opinion 1} \newline \emph{participant} & It depends on the allocation of governmental funding to all other areas of life. Priorities need to be in place, with the likes of the NHS being an example of a high priority area. Housing is an important aspect of life, so it does make sense that the government should subsidize house prices if the budget allows for it in the grand scheme of how the budget is being used. \\
    \midrule
    \textbf{\normalsize Opinion 2} \newline \emph{participant} &Yes, the government should subsidize house prices in so far as it the government's policy to provide universal hosing. Doing so would help to ease the cost of living crisis that is facing the country at the moment. However, there should be a prerequisite for getting such assistance including being in full employment and a minimum pay threshold. \\
    \midrule
    \textbf{\normalsize Opinion 3} \newline \emph{participant} & There are so many struggling to get on the property ladder that subsidizing the cost of house prices would be very welcome and help a lot of people.  Subsidized housing should not be given to anyone and they must meet certain criteria for example being in employment and not claiming benefits.  I wouldn't want this to be seen as a handout to the idle.  There are genuine people out there that require help and I would want them to benefit. \\
    \midrule
    \textbf{\normalsize Opinion 4} \newline \emph{participant} & No. House prices will always find their own level and any interference in this is unnecessary. If the government start interfering in this it will cause many problems. People will expect help in every aspect of life. It will not give people the incentive to work hard and save. Owning your own house has long been seen as a reward for hard work and careful saving. \\
    \midrule
    \textbf{\normalsize Opinion 5} \newline \emph{participant} & If minimum and living wages remain so low in comparison to house prices then yes, the government should subsidize house prices. Originally you were able to buy a house/flat with a little saving and your wage. Now it is almost impossible to buy on your own. Rent is extortionate and so saving is harder whilst you're waiting to buy. Now the cost of living has also gone up making it even harder to save for that deposit too. \\ 
    \midrule \midrule
    \textbf{\normalsize Consensus} \newline \emph{SFT-Utilitarian 70B} & We believe that the government should subsidize house prices. There are so many struggling to get on the property ladder that subsidizing the cost of house prices would be very welcome and help a lot of people. Subsidized housing should not be given to anyone and they must meet certain criteria for example being in employment and not claiming benefits. I wouldn't want this to be seen as a handout to the idle. There are genuine people out there that require help and I would want them to benefit.\\
    \midrule
    \textbf{\normalsize Consensus} \newline \emph{Few-shot 70B} & The government should subsidize house prices in so far as it is the government's policy to provide universal housing. Doing so would help to ease the cost of living crisis that is facing the country at the moment. However, there should be a prerequisite for getting such assistance including being in full employment and a minimum pay threshold.\\
    \midrule
    \textbf{\normalsize Consensus} \newline \emph{Zero-shot 70B} & No. The government should not subsidize house prices.\\
    \midrule
    \textbf{\normalsize Consensus} \newline \emph{SFT-Utilitarian 1B} & Yes, the government should subsidize house prices. This would help people on low incomes get on the property ladder. It would also help the housing market and reduce the amount of empty properties in the UK. However, the government should ensure that people are in full employment and not claiming benefits before they are eligible for this assistance.\\
    \midrule
    \textbf{\normalsize Consensus} \newline \emph{Few-shot 1B} & Consensus Statement: The jury felt that the government should subsidize house prices in so far as it is seen to be in the national interest.\\
    \midrule
    \textbf{\normalsize Consensus} \newline \emph{Zero-shot 1B} & 	The above consensus result highlights a very high degree of acceptance of the government's current policies. This is something that can be seen as an accomplishment for the prime minister and his administration. A level of agreement with the policy can be seen as being a high level of political support for the current policy. It also shows a fairly high degree of acceptance of the policies that are currently in place.\\
    \bottomrule
    \end{tabular}
\end{table}

\subsection{Win-rate analysis on group median agreement scores}

\begin{figure}
    \centering
    \includegraphics[width = 0.66\linewidth]{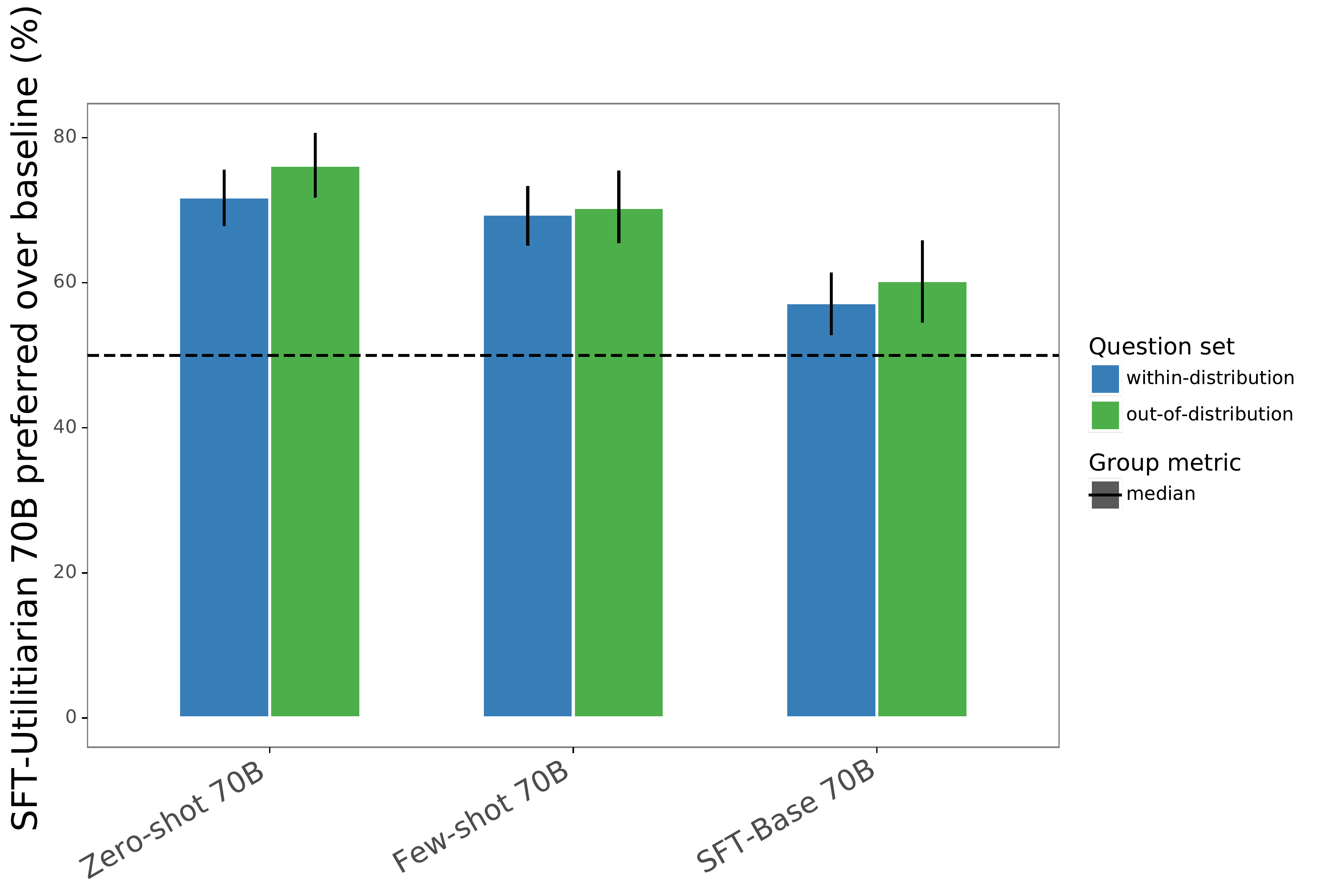}
    \caption{Win rates for comparing models constructed by pairwise comparison of Likert agreement ratings (excluding ties), aggregating scores of a group by taking the median, for within-distribution (blue) and out-of-distribution (green) question sets. Error-bars represent 95\% bootstrapped confidence intervals.}
    \label{fig:median}
\end{figure}

Here, we compare how likely a group is to prefer one policy over another by examining the median agreement scores for each group.
The median aggregation function is a complementary statistic to analyze (in addition to the mean) for our data of Likert ratings, where we expect the resulting distributions to be non-Gaussian (e.g., for our divisive questions, we expect bi-modal distributions). 
Further, for small sample sizes, the median is more robust to outliers.
The substantive conclusions for our primary set of analyses are unchanged by examining the median (Figure \ref{fig:median}).

\subsection{Regression analysis details}

In the main text, we report the results of a mixed-effects logistic generalized linear regression model.
This mixed-model allows us to take into account the fact that the individual ratings are not i.i.d., and instead structured in such a way that multiple ratings come from the same participant and that each question receives multiple ratings.
We constructed a maximal random-effects model, with by-participant and by-question effects of intercept and slope (i.e., effect of model variant), which is the standard in confirmatory hypothesis testing \cite{barr2013random}. In lmer-style syntax, the model formula was \texttt{Likert\_rating $\sim$ model\_variant + (1 + model\_variant | participant) + (1 + model\_variant | question)}.
These random effects capture the idea that our various models might be better on some questions than on other and that different participants might show different levels of preference for some model variants over others.
The model is additionally an \textit{ordinal} regression model. 
The virtue of ordinal regression here is that it preserves the meaning of the points along the Agreement scale, while not stipulating the distance between the points is uniform \cite{liddell2018analyzing}.

We ran a parallel set of analyses for the out-of-distribution evaluation set, which we report here. 
Consistent with our win-rate analysis, we found that participants agreed more strongly with the consensus statements generated by the \emph{SFT-Utilitarian model} than those generated by the \emph{SFT-Base} ($\beta = 0.26$; SE = $0.071$; z = $3.6$; $p = 0.0002$). Participants also preferred the \emph{SFT-Base}'s generations more so than those of the \emph{few-shot} model ($\beta = 0.35$; SE = $0.071$; z = $4.9$; $p < 0.001$), but, as before in the within-distribution evaluation, exhibited no preference for the 
\textit{few-shot} model over the \textit{zero-shot} model ($\beta = 0.17$; SE = $0.089$; z = $1.4$; $p = 0.16$).

\subsection{Comparison of reranking with different Social Welfare Functions}

\begin{figure}[b]
    \centering
    \includegraphics[width = 0.9\linewidth]{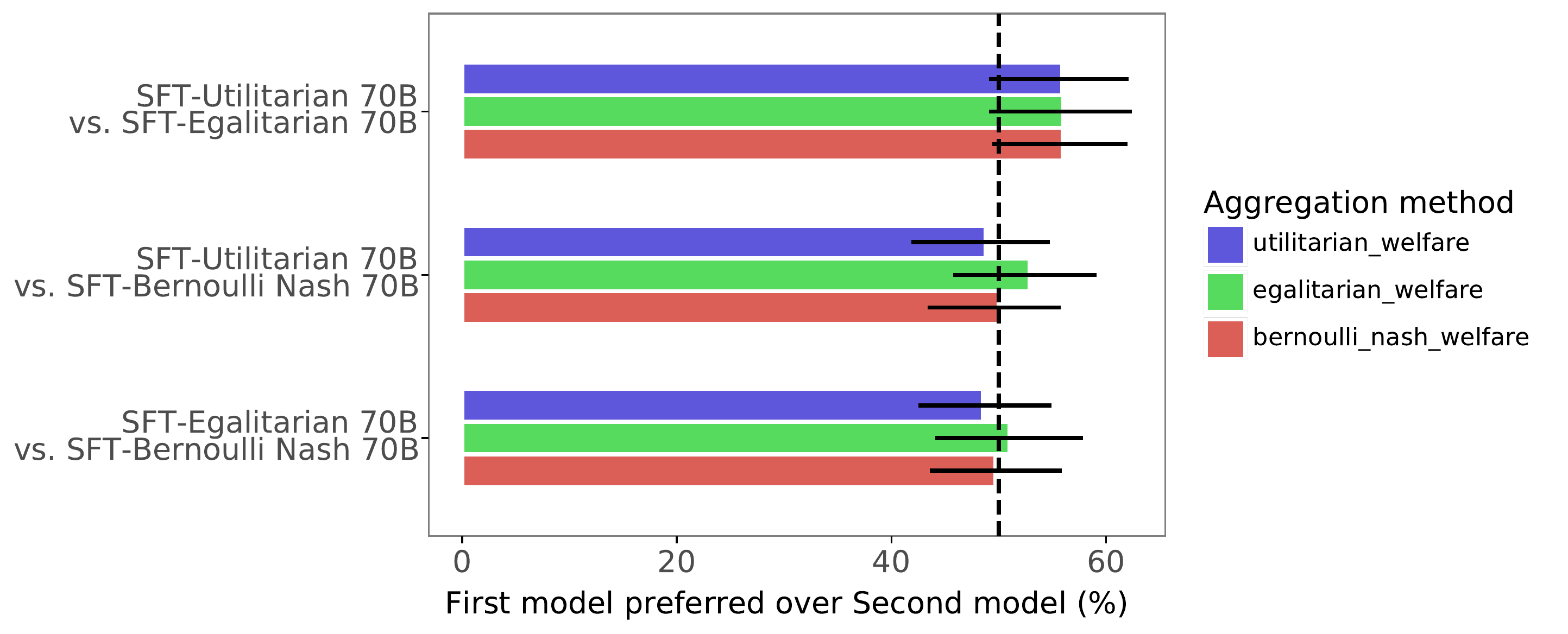}
    \caption{Win rates for comparing SFT models using different aggregation functions in the re-ranking step. Win rates constructed by pairwise comparison of Likert agreement ratings (excluding ties). Control over re-ranking step did not show predicted effects on the welfare of the group. Error-bars represent 95\% bootstrapped confidence intervals. }
    \label{fig:sft}
\end{figure}

Our reward model re-ranking explicitly allows us to specify different social welfare functions to use when aggregating the set of participant-specific reward model predictions. We operationalized this flexibility by considering a parameterised family of social welfare functions described by Equation 1 (main text). A single parameter allows us to select either a pure Utilitarian aggregation function (outputs the mean score across the group), a pure Egalitarian function (outputs the minimum score), or a hybrid prioritizational function such as the Bernoulli-Nash function (the product of the scores), or any other function falling between these extremes. During training, we selected candidates by using a lognormal sampling scheme in order to ensure that the models were trained with data spanning the full distribution of social welfare functions. In order to assess the degree to which our trained models were sensitive to the specific social welfare functions, we ran a human experiment where we used our main model (SFT with reranking) to generate candidates under three distinct welfare functions:
\begin{itemize}
    \item Utilitarian
    \item Bernoulli-Nash
    \item Egalitarian (or Rawlsian)
\end{itemize}

To analyse this data, for each group we computed the welfare score from the set of participant agreement ratings, under each of our three welfare functions of interest. If our approach were indeed sensitive to the aggregation function deployed during re-ranking, we would expect to find that the candidate selected under the Utilitarian function ought to have a higher Utilitarian welfare score than either the Egalitarian or Bernoulli-Nash candidates. A similar result should apply under each of the other welfare functions, such that the candidate selected under that method should show the highest welfare score compared to the other candidates. The pairwise win-rates were computed for each pair of candidates and welfare functions, and as shown in Figure~\ref{fig:sft}, none of the candidates showed the expected welfare score advantage. Indeed, we note that the only pairwise comparison to show a confidence interval that does not include chance is a comparison of the Utilitarian vs. Egalitarian candidates, where contrary to our expectations the Utilitarian candidate actually shows a higher Egalitarian welfare score. This result dovetails with our main paper results where we find a significant boost in minimum agreement scores even under the Utilitarian welfare function. Overall, we believe this null result may be at least partially driven by the fact that even the model using a utilitarian welfare function appears to result in a prioritarian boost in agreement scores, making it highly correlated in expected outcome with the other welfare functions. 

\subsection{Ratings on Position Statements and Question Divisiveness}

\begin{figure}
     \centering
     \begin{subfigure}[b]{0.85\textwidth}
         \centering
         \includegraphics[width=\textwidth]{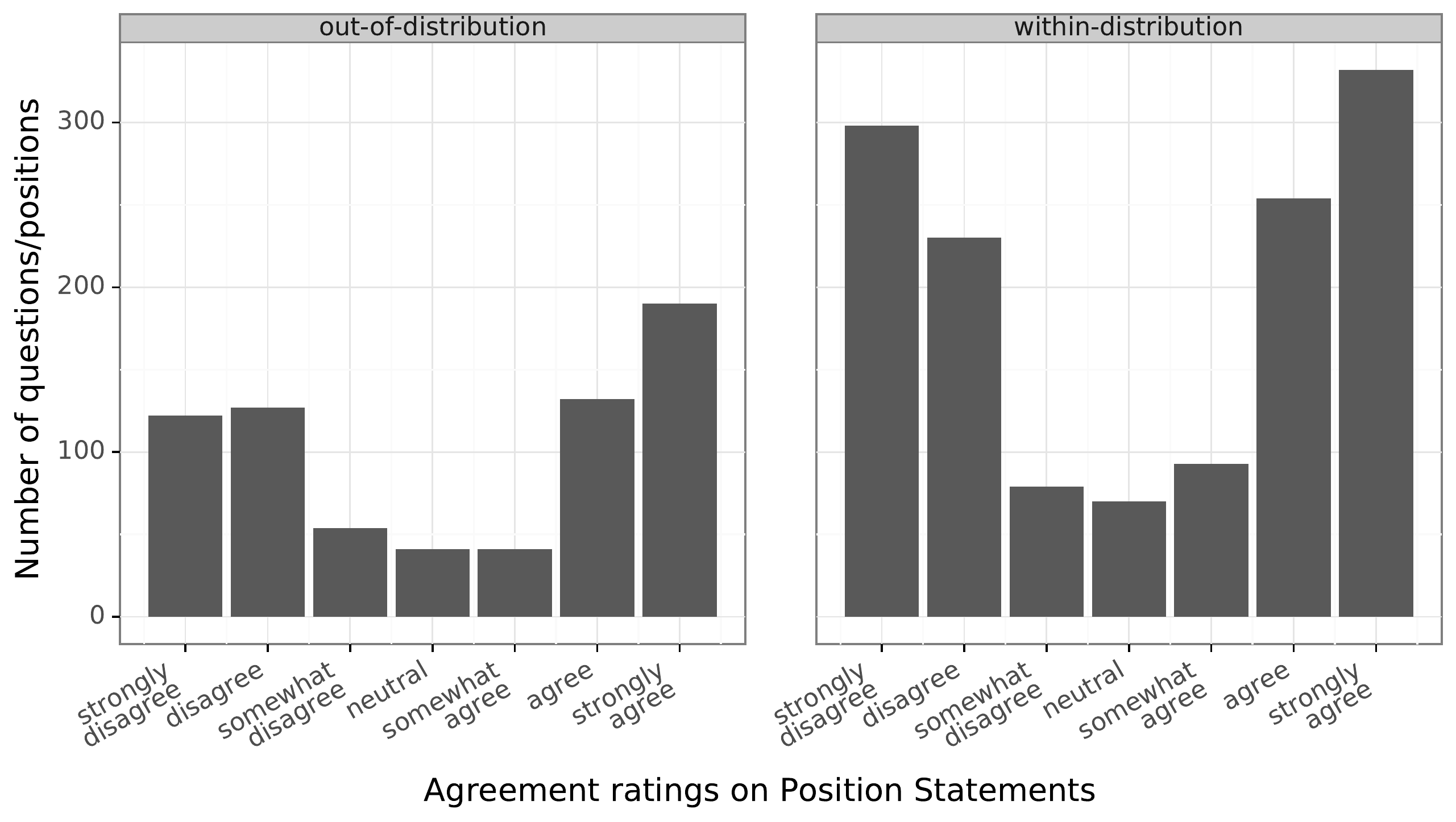}
         \caption{Agreement ratings for Position Statements collapsed across groups and questions for within-distribution and out-of-distribution evaluation data sets.}
         \label{fig:ps_agree}
     \end{subfigure}
     
     \begin{subfigure}[b]{0.85\textwidth}
         \centering
         \includegraphics[width=\textwidth]{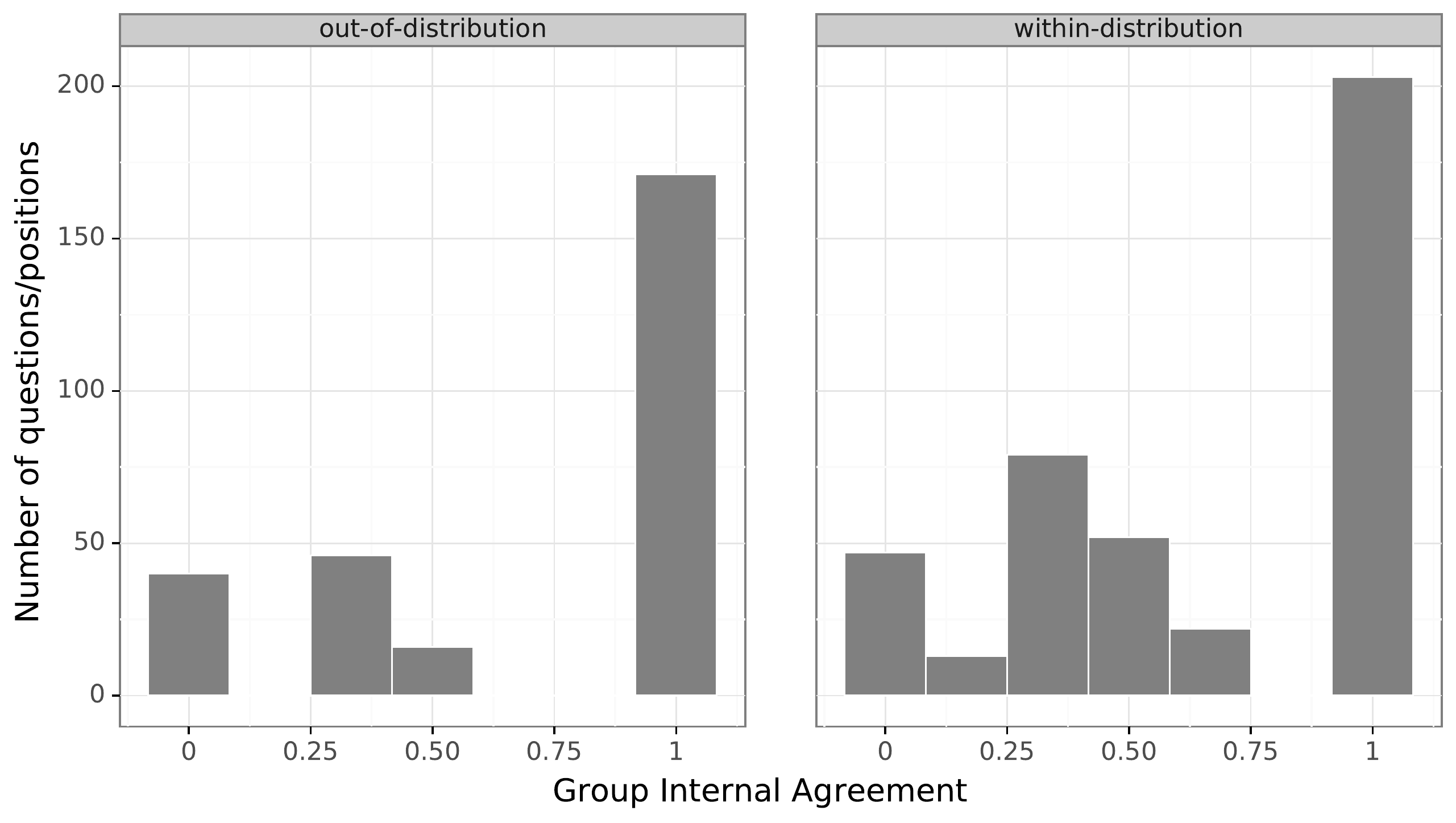}
         \caption{Group Internal Agreement (see text for definition) for Position Statements collapsed across questions for within-distribution and out-of-distribution evaluation data sets.}
         \label{fig:ps_internal}
     \end{subfigure}
     \caption{Descriptive statistics of agreement ratings for positions statements.}
     \label{fig:position_statements}
\end{figure}

Before providing their own opinions, participants provided an agreement rating on the Position Statement, which is a version of the question stated declaratively (e.g., ``The government should increase taxes on the rich.''). 
Overall, participants had relatively strong opinions about the Position Statements, with only about 5\% (in both within-distribution and out-of-distribution evaluation sets) of participant ratings reporting a rating on the midpoint (``neutral'') of the scale (Figure \ref{fig:ps_agree}).

To assess the diversity of our groups with respect to their opinions on the topics under discussion, we computed a Group Internal Agreement score for each question addressed by each group. Group Internal Agreement binarizes participant ratings and normalizes the proportion of the group that agrees with the position statement (removing midpoint ratings) so that agreement and disagreement are treated symmetrically. 

$$
\textrm{Group Internal Agreement} = \left(\frac{1}{2} - \frac{n(r > 4)}{n(r > 4) + n(r < 4)}\right) \times 2    
$$

where $n(r > 4)$ is the number of agreement ratings of 5, 6, or 7 (``somewhat agree'', ``agree'', ``strongly agree'') and $n(r < 4)$ is the number of agreement ratings of 3, 2, or 1 (``somewhat disagree'', ``disagree'', ``strongly disagree''). A Group Internal Agreement score of 1 represents total internal agreement such that all members of the group either (at least somewhat) agree or all (at least somewhat) disagree with the position statement. A score of 0 reflects maximal disagreement in the group such that half of members (at least somewhat) agree and half of members (at least somewhat) disagree with the position statement. Note that an exact score of 0 is only possible when the number of ratings excluding neutral ratings is even.
We find that 50.6\% of questions addressed by groups in the within-distribution evaluation had an Internal Agreement score less than 1 (i.e., there was at least one dissenter) and that 37.1\% in the out-of-distribution evaluation had an Internal Agreement score less than 1 (Figure \ref{fig:ps_internal}). We use a Group Internal Agreement score of less than 1 to denote ``divisive'' questions in the main text, whereas questions with a Group Internal Agreement of 1 are ``undivisive'' questions.

\begin{figure}
    \centering
    \includegraphics[width=\textwidth]{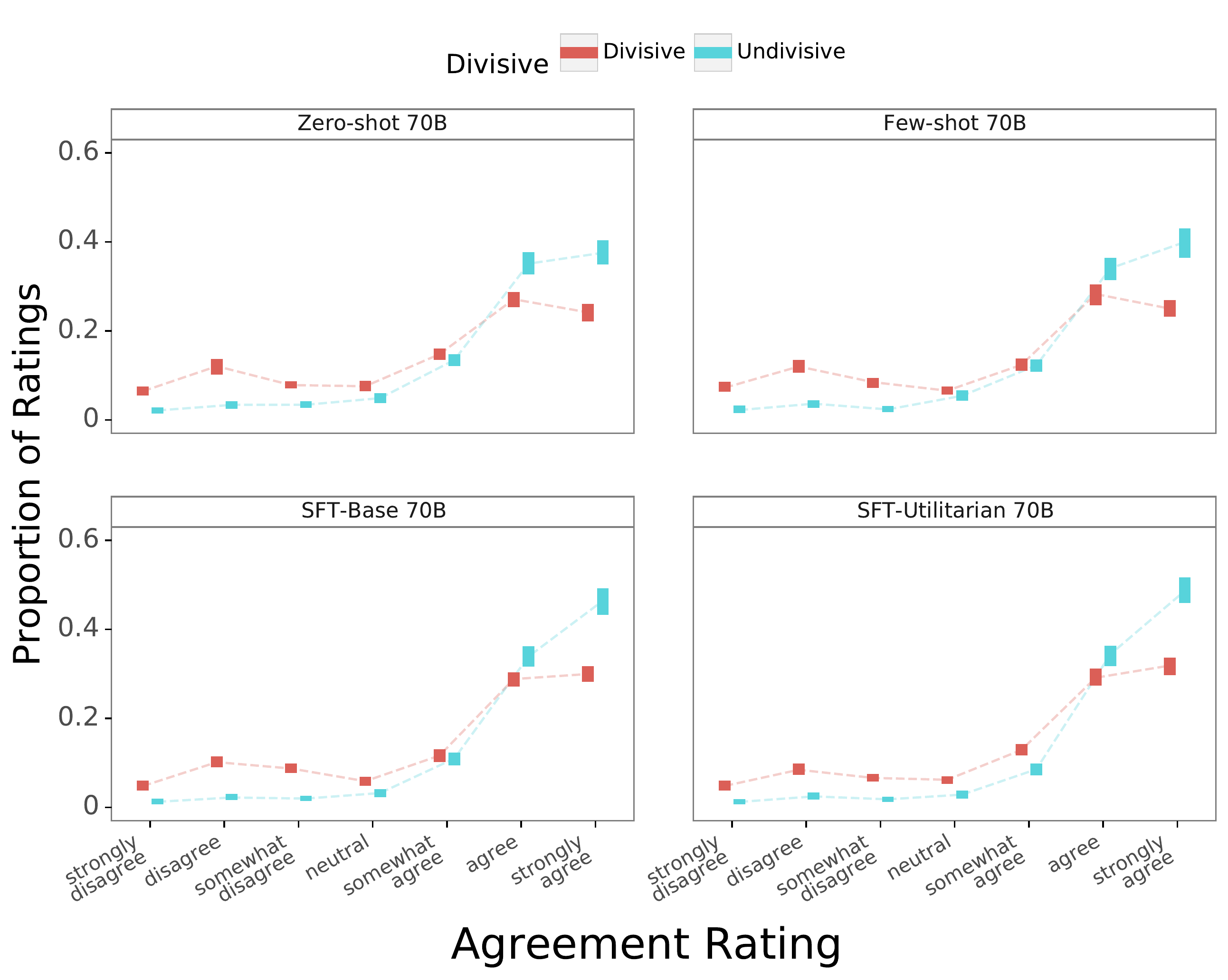}
    \caption{A comparison of 70B model agreement ratings for divisive and undivisive questions. Error-bars represent 95\% bootstrapped confidence intervals.}
    \label{fig:likerts_divisive}
\end{figure}

\subsection{Divisiveness of Candidate Consensus statements}

To assess the performance of the models in more absolute terms, we compared the divisiveness of the candidate consensus statements to the divisiveness of the initial position statements. For the initial position statements, divisiveness is an unsigned measure of Group Internal Agreement (e.g., a position statement to which all participants agreed is treated the same as a position statement to which all participants disagreed). For candidate consensus statements, divisiveness is a signed measure: It is the proportion of participants who agree (in a binarized fashion, as above) with the statement. The term \emph{consent} may thus be more appropriate for this measure. We maintain the term Candidate Consensus Divisiveness to highlight the continuity in the mathematical definition with the Position Statement Divisiveness. 

$$
\textrm{Candidate Consensus Divisiveness} = \frac{n(r > 4)}{n(r > 4) + n(r < 4)}
$$

We first examined the proportion of candidates that were less divisive than the Position Statements. We only consider the Divisive questions (i.e., those questions for which there was some internal disagreement on, approximately 50\% of questions) as it is impossible for a candidate to be less divisive than an Undivisive Position Statement. We find that 65.6\% [61.9, 69.3] of candidates generated from the SFT-Utilitarian model were less divisive than the corresponding Position Statements (the SFT-Base model achieves 58.8\% [54.9, 62.6], the Few-shot prompted model achieves 53.1\% [49.4, 57.2] and the Zero-shot prompted model achieves 54.6\% [50.8, 58.4] improvements upon the position statement divisiveness).

We next examined on each round whether or not a candidate from each of the baseline models achieves \emph{unanimous consent} (i.e., all participant at least ``somewhat agrees'' with a candidate consensus). Again, we look specifically at the divisive questions. 
Under this analysis, we find that the SFT-Utilitarian model generates a candidate that achieves unanimous consent on 40.8\% [35.4, 46.2] of rounds, the SFT-Base model achieves unanimous consent on 31.9\% [26.8, 36.9] of rounds, the Few-shot model achieves unanimous consent on 31.6\% [26.4, 36.6] of rounds, and the Zero-shot prompted model achieves unanimous consent on 33.8\% [28.7, 39.2] of rounds. Together, these results suggest that even in cases that one might expect it to be difficult to find consensus, our model tends to find some common ground.

\end{document}